\documentclass[journal]{IEEEtran}
\usepackage{blindtext}
\usepackage{graphicx}
\usepackage{array}
\usepackage[numbers,sort&compress]{natbib}
\usepackage{amsmath}
\usepackage{color}
\usepackage{algorithm2e}
\usepackage{subcaption}
\let\labelindent\relax
\usepackage{enumitem}
\usepackage{url}
\usepackage{multirow}
\usepackage{moresize}
\usepackage[table]{xcolor}
\usepackage{color,soul}
\usepackage[]{hyperref}
\soulregister\cite7
\soulregister\ref7
\soulregister\pageref7
\usepackage{comment}
\usepackage{dblfloatfix}    
\usepackage[flushleft]{threeparttable} 
\usepackage{enumitem}
\usepackage{xcolor}
\usepackage[geometry]{ifsym}
\usepackage{bm}

\begin{document}

\twocolumn[{%
\vspace{40mm}
{ \large
\begin{itemize}[leftmargin=2.5cm, align=parleft, labelsep=2.0cm, itemsep=4ex,]

\item[\textbf{Citation}]{D. Temel, M-H. Chen, and G. AlRegib, "Traffic Sign Detection under Challenging
Conditions: A Deeper Look Into Performance
Variations and Spectral Characteristics," IEEE Transactions on Intelligent Transportation Systems, 2019.}

\item[\textbf{Dataset}]{\url{https://github.com/olivesgatech/CURE-TSD}}

\item[\textbf{Bib}]  {@INPROCEEDINGS\{Temel2019\_ITS,\\ 
author=\{D. Temel and M.-H. Chen and G. AlRegib\},\\ 
booktitle=\{IEEE Transactions on Intelligent Transportation Systems\},\\
title=\{Traffic Sign Detection under Challenging Conditions: A Deeper Look Into Performance Variations and Spectral Characteristics\},\\ 
year=\{2019\}, \} 
}

\item[\textbf{DOI}]{10.1109/TITS.2019.2931429}

\item[\textbf{ISSN}]{1524-9050}

\item[\textbf{Contact}]{\href{mailto:alregib@gatech.edu}{alregib@gatech.edu}~~~~~~~\url{https://ghassanalregib.info/}\\ \href{mailto:dcantemel@gmail.com}{dcantemel@gmail.com}~~~~~~~\url{http://cantemel.com/}}
\end{itemize}
\thispagestyle{empty}
\newpage
\clearpage
\setcounter{page}{1}
}
}]

\title{Traffic Sign Detection under Challenging Conditions: A Deeper Look Into Performance Variations and Spectral Characteristics}

\author{Dogancan~Temel,~Min-Hung~Chen,~and~Ghassan~AlRegib

{\includegraphics[width=0.19\linewidth]{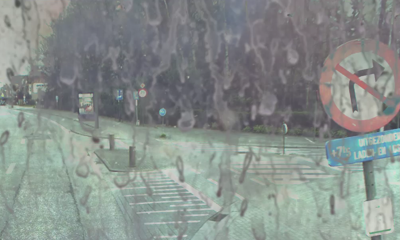}}
{\includegraphics[width=0.19\linewidth]{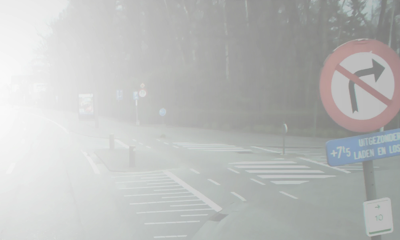}}
{\includegraphics[width=0.19\linewidth]{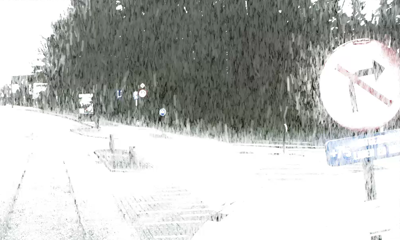}} 
{\includegraphics[width=0.19\linewidth]{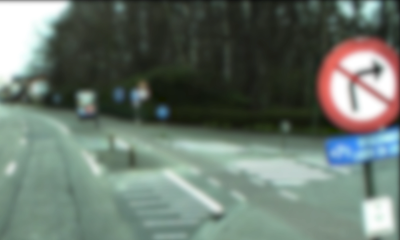}}
{\includegraphics[width=0.19\linewidth]{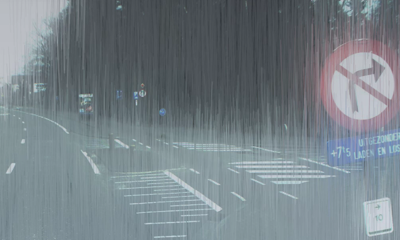}}
\captionof{figure}{Challenging scene examples from the \texttt{CURE-TSD-Real} dataset: dirty lens, haze, snow, blur, and rain, respectively.}
\label{fig:sample_scenes}
\vspace{-10mm}

\thanks{Authors are with the School of Electrical and Computer Engineering, Georgia Institute of Technology, Atlanta, GA, 30332-0250 USA. Corresponding author email: cantemel@gatech.edu. }
}

\markboth{}%
{Shell \MakeLowercase{\textit{et al.}}: Bare Demo of IEEEtran.cls for Journals}

\maketitle

\begin{abstract} 
Traffic signs are critical for maintaining the safety and efficiency of our roads. Therefore, we need to carefully assess the capabilities and limitations of automated traffic sign detection systems. Existing traffic sign datasets are limited in terms of type and severity of challenging conditions. Metadata corresponding to these conditions are unavailable and it is not possible to investigate the effect of a single factor because of simultaneous changes in numerous conditions. To overcome the shortcomings in existing datasets, we
introduced the CURE-TSD-Real dataset, which is based on simulated challenging conditions that correspond to adversaries that can occur in real-world environments and systems. We test the performance of two benchmark algorithms and show that severe conditions can result in an average performance degradation of $29\%$ in precision and $68\%$ in recall. We investigate the effect of challenging conditions through spectral analysis and show that challenging conditions can lead to distinct magnitude spectrum characteristics. Moreover, we show that mean magnitude spectrum of changes in video sequences under challenging conditions can be an indicator of detection performance. CURE-TSD-Real dataset is available online at \href{https://github.com/olivesgatech/CURE-TSD}{https://github.com/olivesgatech/CURE-TSD.}

\end{abstract}


\IEEEpeerreviewmaketitle

\section{Introduction}
Transportation systems are transformed by disruptive technologies that are based on autonomous systems. In order for autonomous systems to seamlessly operate in real-world conditions, they need to be robust under challenging conditions. In this study, we focus on automated traffic sign detection systems and investigate the effect of challenging conditions in algorithmic performance. Currently, the performance of traffic sign detection algorithms are tested with existing traffic sign datasets in the literature \cite{Grigorescu2003,Timofte2009,Timofte2014,Belaroussi2010,Larsson2011,Stallkamp2011,Stallkamp2012,Houben2013,Mogelmose2012,Zhu2016,Yang2016,Zhang2017}, which have been very useful to develop and evaluate state-of-the-art traffic sign recognition and detection algorithms. However, these datasets are usually very limited in terms of type and severity of challenging conditions. Moreover, there is no metadata corresponding to the types and the levels of these conditions. Traffic sign size can be considered as the only metadata corresponding to the challenge level, which can degrade the detection performance significantly \cite{Yi2018}. Besides the limitations of challenging conditions, it is not feasible to assess the effect of most of the challenging conditions in existing datasets because of limited control over the acquisition process. A number of conditions change simultaneously, which makes it impossible to investigate the effect of a specific condition. To overcome the shortcomings of existing datasets and enable assessing the effect of challenging conditions one at a time, we introduced traffic sign recognition \cite{Temel2017_NIPSW} and detection datasets \cite{Temel2018_SPM,Temel2019}. We hosted the first IEEE Video and Image Processing (VIP) Cup \cite{Temel2018_SPM} within the IEEE Signal Processing Society and obtained an algorithmic benchmark for the CURE-TSD dataset, which is based on video sequences corresponding to real-world as well as simulator environments. In this study, we focus on the real-world sequences denoted as \texttt{CURE-TSD-Real} dataset. Specifically, we investigate the effect of challenging conditions over the average performance of top two benchmark algorithms that are based on deep neural networks.

Recently, adversarial examples have been commonly used in the literature to test the vulnerability of recognition and detection systems \cite{Szegedy2014b,Goodfellow2015}. Even though adversarial examples have been successful in deceiving these generic object recognition and detection systems, they have not been effective in traffic sign recognition and detection systems \cite{Lu2017, Das2017}. Lu \textit{et al.} \cite{Luo2017} showed that adversarial examples deceive traffic sign detection systems only in limited scenarios and Das \textit{et al.} \cite{Das2017} showed that a simple compression stage can minimize the effect of adversarial attacks in traffic sign recognition. Adversarial examples are useful to assess the limits of existing systems with special inputs that are optimized for deception. However, such adversarial examples do not necessarily correspond to real-world challenging conditions. Moreover, previous studies mainly focused on feeding adversarial data directly into the classification model. However, in real world, challenging conditions can directly affect the data acquired by sensors rather than classifiers. In this study, we differentiate from previous studies by focusing on challenging conditions corresponding to adversaries that can naturally occur in real-world environments and systems as shown in Fig.~\ref{fig:sample_scenes}. On contrary to adversarial studies in the literature that require model information to design input images, we designed challenging conditions independent of the detection algorithms, which enables a black-box performance assessment. Previously, we performed black-box assessment of object detection APIs with realistic challenging conditions in \cite{Temel2018_CUREOR, Temel2019_CUREOR}, and investigated the robustness and out-of-distribution classification performance of traffic sign classifiers in \cite{Prabhushankar2018,Kwon2019}.     

In addition to investigating the effect of challenging conditions in traffic sign detection performance, we also analyze the effect of challenging conditions in terms of spectral characteristics. In \cite{VANDERSCHAAF1996}, Van der Schaaf and Van Hateren  analyzed the power spectrum of natural images and showed that there is a common characteristic followed by natural images. In \cite{Torralba2003}, Torralba and Olivia extracted more information based on spectrum related to the openness of images, the semantic category of scenes, the recognition of objects, and the depth of scenes. Based on these observations and findings, a direct comparison between spectrum of challenge-free sequences and challenging sequences can be affected by the context of the scene. However, if we first obtain the difference between challenge-free and challenging sequences and then obtain the power spectrum, we can limit the effect of context and concentrate on the change with respect to reference conditions, which is the methodology pursued in this study.

The rest of this paper is organized as follows: In Section \ref{sec:related}, we analyze existing traffic sign datasets. In Section \ref{sec:cure}, we provide a general description of the \texttt{CURE-TSD-Real} dataset, describe challenging conditions, and briefly introduce benchmark algorithms. We discuss traffic sign detection performance under challenging conditions and analyze spectral characteristics of these conditions in Section \ref{sec:perf}. Specifically, we explain the performance metrics in Section \ref{subsec:perf_metrics}, describe the training and test datsets in Section \ref{subsec:perf_data},  report performance variation under challenging conditions in Section \ref{subsec:perf_overall}, perform a spectral analysis of challenging conditions in Section \ref{subsec:perf_spectrum}, and analyze the relationship between detection performance and spectral characteristics in Section \ref{subsec:perf_estimation}. Finally, we conclude our work in Section \ref{sec:conc}.

\begin{center}
\begin{table*}[htbp!]
\centering
\caption{Main characteristics of publicly available datasets and CURE-TSD-Real dataset* }
\label{tab_datasets}

\begin{threeparttable}

\begin{tabular}{cccccccccc}
\hline

 &\textbf{\begin{tabular}[c]{@{}c@{}}Number of\\videos\end{tabular}}  &\textbf{\begin{tabular}[c]{@{}c@{}}Number of\\annotated\\ images\end{tabular}}   &\textbf{\begin{tabular}[c]{@{}c@{}}Number of \\ annotated \\signs\end{tabular}}  & \textbf{\begin{tabular}[p]{@{}c@{}}Annotation\\ information\end{tabular}}  &\textbf{Resolution}   &
\textbf{\begin{tabular}[c]{@{}c@{}}Number of \\ sign types\end{tabular}}  & \textbf{\begin{tabular}[c]{@{}c@{}}Annotated \\ sign size\end{tabular}}    & \textbf{\begin{tabular}[c]{@{}c@{}}Acquisition \\ location \end{tabular}}   & 
\textbf{\begin{tabular}[c]{@{}c@{}}Challenging \\ conditions \\ (*=annotated) \end{tabular}}       
 \\ \hline


\textbf{RUG  \cite{Grigorescu2003}}   &N/A  &N/A  &N/A & N/A &360X270  &3  &N/A  &Netherlands & \begin{tabular}[c]{@{}c@{}}illumination\\ overcast\\ blur \end{tabular}  \\ \hline

\textbf{\href{http://btsd.ethz.ch/shareddata/}{BelgiumTS} \cite{Timofte2009}}  &4  &9,006 &13,444 &\begin{tabular}[c]{@{}c@{}}sign type \\ bounding box \\3D location\end{tabular}  &1,628x1,236  &62  & \begin{tabular}[c]{@{}c@{}}9x10 to\\ 206x277\end{tabular} &Belgium & \begin{tabular}[c]{@{}c@{}}illumination\\occlusion\\ overcast \end{tabular}   \\ \hline

\textbf{\href{http://btsd.ethz.ch/shareddata/}{BelgiumTSC} \cite{Timofte2014}}  &N/A  &7,125  &7,125 &\begin{tabular}[c]{@{}c@{}}sign type \\ bounding box \end{tabular}  &  \begin{tabular}[c]{@{}c@{}}22x21 to \\ 674x527\end{tabular} &62  &\begin{tabular}[c]{@{}c@{}}11x10 to\\ 562x438\end{tabular}  &Belgium & \begin{tabular}[c]{@{}c@{}} illumination\\occlusion\\deformation  \end{tabular} \\ \hline

\textbf{\href{http://www.itowns.fr/roadsign.php}{Stereopolis} \cite{Belaroussi2010}} &N/A  &273  &273 &\begin{tabular}[c]{@{}c@{}}sign type \\ bounding box \end{tabular}  &1,920x1,080  &10  & \begin{tabular}[c]{@{}c@{}}25x25 to\\ 204x159 \end{tabular} &France &\begin{tabular}[c]{@{}c@{}}illumination\\occlusion\\overcast\end{tabular}  \\ \hline

\textbf{\href{http://www.cvl.isy.liu.se/en/research/datasets/traffic-signs-dataset/}{STS} \cite{Larsson2011}}  &N/A  &3,488  &3,488 &\begin{tabular}[c]{@{}c@{}}sign type  \\ bounding box\\ visibility status \\ road status\end{tabular}  &1,280x960 &7  & \begin{tabular}[c]{@{}c@{}}3x5 to\\ 263x248\end{tabular}  &Sweden & \begin{tabular}[c]{@{}c@{}}illumination\\occlusion*\\ shadow, blur*\\overcast, rain
\end{tabular}  \\ \hline

\textbf{\href{http://benchmark.ini.rub.de/}{GTSRB} \cite{Stallkamp2011,Stallkamp2012}}   &N/A  &51,840  &51,840 &sign type  & \begin{tabular}[c]{@{}c@{}}15x15 to \\ 250x250\end{tabular} &43  & \begin{tabular}[c]{@{}c@{}}15x15 to \\ 250x250\end{tabular} &Germany & \begin{tabular}[c]{@{}c@{}} illumination\\occlusion, blur\\deformation \\perspective   \end{tabular} 
  \\ \hline

\textbf{\href{http://cvrr.ucsd.edu/LISA/lisa-traffic-sign-dataset.html}{LISA}  \cite{Mogelmose2012}}  &17 tracks  &6,610  &7,855 &\begin{tabular}[c]{@{}c@{}}sign type  \\bounding box\\occlusion status\\ road status\end{tabular}  & \begin{tabular}[c]{@{}c@{}}640x480 to\\ 1,024x522\end{tabular} &47  & \begin{tabular}[c]{@{}c@{}}6x6 to \\ 167x168\end{tabular} &USA &\begin{tabular}[c]{@{}c@{}} illumination\\occlusion*\\ shadow, blur\\ reflection\\codec error\\dirty lens\\ decolorization* \\overcast  \end{tabular}  \\ \hline

\textbf{\href{http://benchmark.ini.rub.de/}{GTSDB} \cite{Houben2013}}  &N/A  &900  &1,206 &\begin{tabular}[c]{@{}c@{}}sign type  \\bounding box\end{tabular}   &1,360X800  &43  &\begin{tabular}[c]{@{}c@{}}16-128  \\longer\\ edge  \end{tabular}  & Germany& \begin{tabular}[c]{@{}c@{}}illumination\\occlusion\\blur, shadow\\haze, rain\\overcast  \end{tabular} \\ \hline

\textbf{\href{http://cg.cs.tsinghua.edu.cn/traffic-sign/}{TT-100K} \cite{Zhu2016} }  &N/A  &100,000  &30,000 & \begin{tabular}[c]{@{}c@{}}sign type  \\bounding box \\ pixel map \end{tabular}  &2,048x2,048  &45  & \begin{tabular}[c]{@{}c@{}}2x7 to \\ 397x394\end{tabular} &China & \begin{tabular}[c]{@{}c@{}}illumination \\occlusion \\overcast\\haze, shadow \end{tabular}   \\ \hline

\textbf{\href{http://luo.hengliang.me/data.htm}{CTSD} \cite{Yang2016}}   &N/A  &1,100  &1,574 &\begin{tabular}[c]{@{}c@{}}sign type  \\bounding box \end{tabular}  & \begin{tabular}[c]{@{}c@{}}1,024x768 \&\\ 1,270x800\end{tabular}  & 48 & \begin{tabular}[c]{@{}c@{}}26x26 to \\ 573x557\\ \end{tabular} &China &  \begin{tabular}[c]{@{}c@{}}illumination\\occlusion\\shadow, rain,\\overcast\\dirty lens\\reflection\\haze, blur  \end{tabular}\\ \hline

\textbf{\href{https://github.com/csust7zhangjm/CCTSDB}{CCTSDB} \cite{Zhang2017}}   &N/A  &10,000  &13,361 & \begin{tabular}[c]{@{}c@{}}sign category  \\bounding box \end{tabular}  & \begin{tabular}[c]{@{}c@{}}1,024x768 \&\\ 1,270x800 \&\\ 1,000x350 \end{tabular}  &3 classes  & \begin{tabular}[c]{@{}c@{}}10x11 to \\ 380x378\\ \end{tabular}  &China & \begin{tabular}[c]{@{}c@{}}illumination\\occlusion\\shadow, rain\\overcast\\dirty lens\\reflection\\haze, blur  \end{tabular} \\ \hline \hline

\textbf{\href{https://github.com/olivesgatech/CURE-TSD}{CURE-TSD-Real}} &2,989  &896,700  &648,186 &\begin{tabular}[c]{@{}c@{}}sign type  \\bounding box \\ challenge type \\ challenge level\end{tabular}   &1,628x1,236  &14  & \begin{tabular}[c]{@{}c@{}} 10x11 to \\ 206x277\end{tabular}  &Belgium & \begin{tabular}[c]{@{}c@{}}rain*, snow* \\ shadow*, haze*\\ illumination* \\decolorization*\\blur*, noise*\\codec error*\\dirty lens*\\occlusion \\overcast \end{tabular}  \\ \hline \hline

\end{tabular}

\begin{tablenotes}
\item[*] Online sources of the datasets are hyperlinked to the dataset names in the first column of this table.
\end{tablenotes}
\end{threeparttable}

\vspace{-5.0mm}

\end{table*}
\end{center}

\section{Existing Datasets}
\label{sec:related}
We summarize the main characteristics of existing traffic sign datasets in Table \ref{tab_datasets}, which includes number of video sequences, number of annotated frames/images, number of annotated signs, annotation information, resolution, number of sign types, annotated sign size, acquisition location, and challenging conditions. When a category does not apply to a specific dataset, there is a `not applicable' abbreviation (N/A). We report the characteristics of publicly available datasets based on the reference papers as well as available dataset files. The majority of listed datasets are based on images whereas LISA provides short tracks up to 30 frames per track and BelgiumTS provides 4 videos whose number of frames varies between 2,001 and 6,001. Total number of annotated images varies from 273 to 100,000 in which number of traffic signs is in between 273 and 51,840. Annotations are mainly based on sign types and bounding box coordinates but 3D location, pixel map, visibility status, occlusion condition, and road status are also provided in certain datasets. The resolution of the traffic sign detection datasets is in between 360x270 and 2,048x2,048 and the sign size in all listed datasets vary from 2x7 to 573x557. The number of traffic sign types vary from 3 to 62. Challenging conditions are not annotated or explicitly described in majority of the datasets. Therefore, we visually inspected these datasets to list apparent challenging conditions, which include illumination, occlusion, shadow, blur, reflection, codec error, dirty lens, overcast, haze, and deformation of the traffic signs. The majority of the listed datasets were captured in Europe but recent studies also include China and USA. All of the datasets include images captured with color cameras but LISA dataset also includes grayscale images directly obtained from car cameras.

\begin{center}
\begin{figure}[htbp!]
\centering
\begin{minipage}[b]{0.18\linewidth}
  \centering
\includegraphics[width=\linewidth,height=\linewidth]{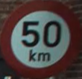}
  \vspace{0.15cm}
  \centerline{\footnotesize{ {\tabular[t]{@{}c@{}}(a) speed\\limit \endtabular} }}
\end{minipage}
\begin{minipage}[b]{0.18\linewidth}
  \centering
\includegraphics[width=\linewidth,height=\linewidth]{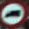}
  \vspace{0.15cm}
  \centerline{\footnotesize{ {\tabular[t]{@{}c@{}}(b) good\\vehicles \endtabular}}}
\end{minipage}
\begin{minipage}[b]{0.18\linewidth}
  \centering
\includegraphics[width=\linewidth,height=\linewidth]{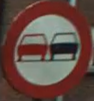}
  \vspace{0.15cm}
  \centerline{\footnotesize{ {\tabular[t]{@{}c@{}}(c) no\\overtaking \endtabular} }}
\end{minipage}
\begin{minipage}[b]{0.18\linewidth}
  \centering
\includegraphics[width=\linewidth,height=\linewidth]{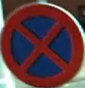}
  \vspace{0.15cm}
  \centerline{\footnotesize{ {\tabular[t]{@{}c@{}}(d) no\\stopping \endtabular} }}
\end{minipage}
\begin{minipage}[b]{0.18\linewidth}
  \centering
\includegraphics[width=\linewidth,height=\linewidth]{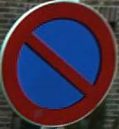}
  \vspace{0.15cm}
  \centerline{\footnotesize{ {\tabular[t]{@{}c@{}}(e) no\\parking \endtabular} }}
\end{minipage}
\begin{minipage}[b]{0.18\linewidth}
  \centering
\includegraphics[width=\linewidth,height=\linewidth]{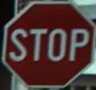}
  \vspace{0.15cm}
  \centerline{\footnotesize{(f) stop }}
\end{minipage}
\begin{minipage}[b]{0.18\linewidth}
  \centering
\includegraphics[width=\linewidth,height=\linewidth]{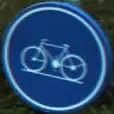}
  \vspace{0.15cm}
  \centerline{\footnotesize{(g) bicycle }}
\end{minipage}
\begin{minipage}[b]{0.18\linewidth}
  \centering
\includegraphics[width=\linewidth,height=\linewidth]{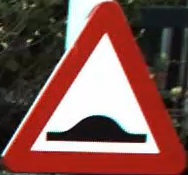}
  \vspace{0.15cm}
  \centerline{\footnotesize{(h) hump }}
\end{minipage}
\begin{minipage}[b]{0.18\linewidth}
  \centering
\includegraphics[width=\linewidth,height=\linewidth]{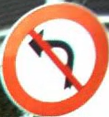}
  \vspace{0.15cm}
  \centerline{\footnotesize{(i) no left }}
\end{minipage}
\begin{minipage}[b]{0.18\linewidth}
  \centering
\includegraphics[width=\linewidth,height=\linewidth]{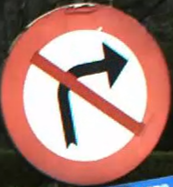}
  \vspace{0.15cm}
  \centerline{\footnotesize{(j) no right }}
\end{minipage}
\begin{minipage}[b]{0.18\linewidth}
  \centering
\includegraphics[width=\linewidth,height=\linewidth]{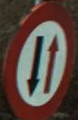}
  \vspace{0.15cm}
  \centerline{\footnotesize{(k) priority to }}
\end{minipage}
\begin{minipage}[b]{0.18\linewidth}
  \centering
\includegraphics[width=\linewidth,height=\linewidth]{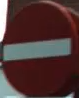}
  \vspace{0.15cm}
  \centerline{\footnotesize{(l) no entry }}
\end{minipage}
\begin{minipage}[b]{0.18\linewidth}
  \centering
\includegraphics[width=\linewidth,height=\linewidth]{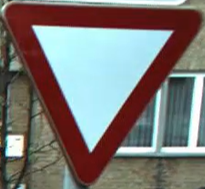}
  \vspace{0.15cm}
  \centerline{\footnotesize{(m) yield }}
\end{minipage}
\begin{minipage}[b]{0.18\linewidth}
  \centering
\includegraphics[width=\linewidth,height=\linewidth]{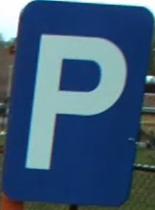}
  \vspace{0.15cm}
  \centerline{\footnotesize{(n) parking }}
\end{minipage}

\vspace{-0.2cm}
\caption{Traffic signs types in the CURE-TSD-Real dataset.}
\label{fig: sign_type}
\vspace{-0.8cm}
\end{figure}
\end{center}

\section{CURE-TSD-Real Dataset}
\label{sec:cure}
\subsection{General Information}
\label{subsec:cure_general}
Among the datasets analyzed in Section~\ref{sec:related}, BelgiumTS \cite{Timofte2009} and LISA \cite{Mogelmose2012} are the only datasets that provide partial video sequences. When this study was conducted, tracks in the LISA dataset were available but not the video sequences. Therefore, we utilized the BelgiumTS \cite{Timofte2009} dataset to obtain video sequences. We selected a subset of the traffic signs in the BelgiumTS dataset as shown in Fig.~\ref{fig: sign_type} and labeled consecutive frames only for these sign types. Sign types were selected according to the compatibility with the synthesized part of the CURE-TSD dataset, which is not considered in this paper. In total, there were four main sequences in BelgiumTS, which included 3,001, 6,201, 2,001, and 4,001 frames. We grouped 300 consecutive frames as individual videos and obtained a total of $49$ videos. In the BelgiumTS dataset, annotations were provided for specific frames and annotating the missing frames could have resulted in an inconsistency among the labels. Therefore, we annotated all the frames including the ones with existing labels and extended the number of annotated frames from 9,006 to 14,700 using the  Video Annotation Tool from Irvine, California (VATIC) \cite{vatic2}. Specifically, we utilized the {\href{https://dbolkensteyn.github.io/vatic.js/}{JavaScript version}} on the browser and labeled a traffic sign if half of the sign was visible. We considered the original video sequences as baseline and added challenging conditions at different levels to test the performance limits of traffic sign detection algorithms.

\begin{center}
\newlength{\tempheight}
\newlength{\tempwidth}
\newlength{\vblank}

\newcommand{\columnname}[1]
{\makebox[\tempwidth][c]{\textbf{#1}}}

\newcommand{\rowname}[1]
{\begin{minipage}[b]{0.5cm}
    \centering\rotatebox{90}{\makebox[\tempheight][c]{\textbf{#1}}}
\end{minipage}}

\begin{figure*}[!htbp]
  \centering

    \setlength{\tempwidth}{.15\linewidth}
    \setlength{\vblank}{0mm}
    \settoheight{\tempheight}{\includegraphics[width=\tempwidth]{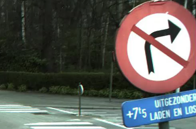}}

    \begin{minipage}[b]{0.5cm}\centering{\makebox[0.5cm][c]{}}\end{minipage}
        \begin{minipage}[c][0.6cm][t]{\tempwidth}\columnname{Level:1}\end{minipage}
        \begin{minipage}[c][0.6cm][t]{\tempwidth}\columnname{Level:3}\end{minipage}
        \begin{minipage}[c][0.6cm][t]{\tempwidth}\columnname{Level:5}\end{minipage}
        \begin{minipage}[c][0.6cm][t]{\tempwidth}\columnname{Level:4}\end{minipage}
        \begin{minipage}[c][0.6cm][t]{\tempwidth}\columnname{Level:5}\end{minipage}

    \rowname{Decolor.}
        {\includegraphics[width=\tempwidth]{./Figs/challenges/01_21_01_01_01_10_0001.png}}\vspace{\vblank}
        {\includegraphics[width=\tempwidth]{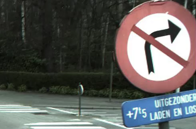}}
        {\includegraphics[width=\tempwidth]{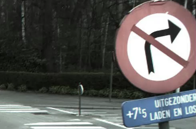}}
        {\includegraphics[width=\tempwidth]{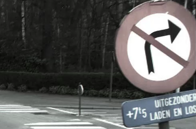}}        
        {\includegraphics[width=\tempwidth]{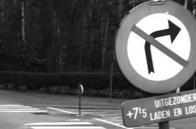}}  \\
    \rowname{Lens blur}
        {\includegraphics[width=\tempwidth]{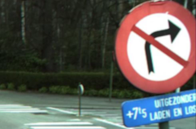}}\vspace{\vblank}
        {\includegraphics[width=\tempwidth]{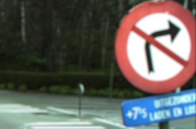}}        
        {\includegraphics[width=\tempwidth]{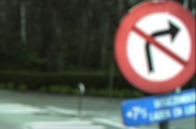}}
        {\includegraphics[width=\tempwidth]{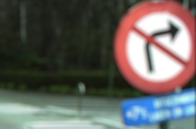}}        
        {\includegraphics[width=\tempwidth]{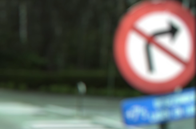}} \\     
        
    \rowname{Codec}
        {\includegraphics[width=\tempwidth]{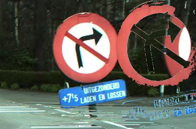}}\vspace{\vblank}
        {\includegraphics[width=\tempwidth]{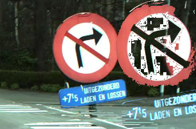}}        
        {\includegraphics[width=\tempwidth]{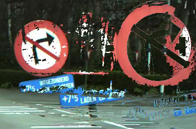}}
        {\includegraphics[width=\tempwidth]{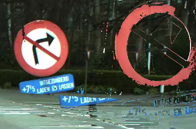}}        
        {\includegraphics[width=\tempwidth]{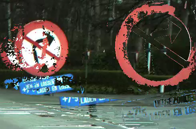}}

    \rowname{Darkening}
        {\includegraphics[width=\tempwidth]{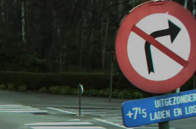}}\vspace{\vblank}
        {\includegraphics[width=\tempwidth]{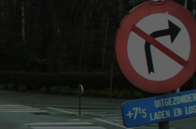}}
        {\includegraphics[width=\tempwidth]{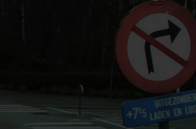}}
        {\includegraphics[width=\tempwidth]{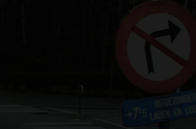}}        
        {\includegraphics[width=\tempwidth]{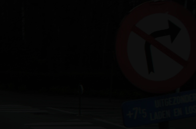}} \\         
    \rowname{Dirty lens}
        {\includegraphics[width=\tempwidth]{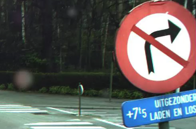}}\vspace{\vblank}
        {\includegraphics[width=\tempwidth]{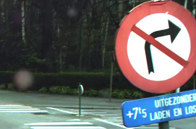}}        
        {\includegraphics[width=\tempwidth]{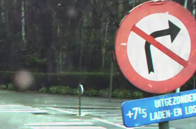}}
        {\includegraphics[width=\tempwidth]{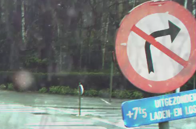}}        
        {\includegraphics[width=\tempwidth]{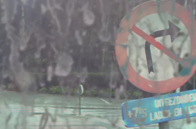}} \\       
    \rowname{Exposure}
        {\includegraphics[width=\tempwidth]{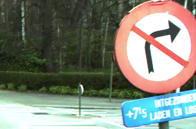}}\vspace{\vblank}
        {\includegraphics[width=\tempwidth]{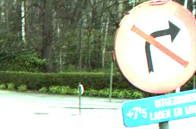}}
        {\includegraphics[width=\tempwidth]{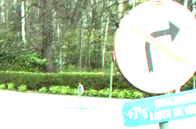}}
        {\includegraphics[width=\tempwidth]{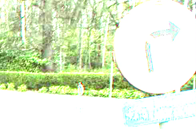}}        
        {\includegraphics[width=\tempwidth]{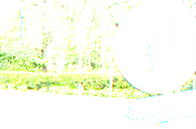}} \\      
    \rowname{Gaussian bl.}
        {\includegraphics[width=\tempwidth]{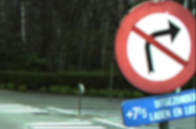}}\vspace{\vblank}
        {\includegraphics[width=\tempwidth]{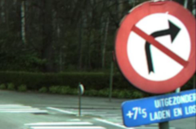}}\vspace{\vblank}        
        {\includegraphics[width=\tempwidth]{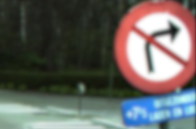}}
        {\includegraphics[width=\tempwidth]{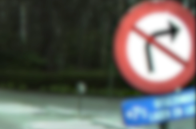}}\vspace{\vblank}        
        {\includegraphics[width=\tempwidth]{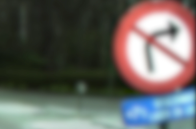}}\\        
    \rowname{Noise}
        {\includegraphics[width=\tempwidth]{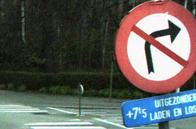}}\vspace{\vblank}
        {\includegraphics[width=\tempwidth]{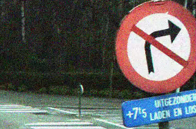}}
        {\includegraphics[width=\tempwidth]{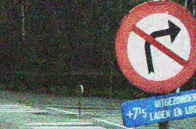}}
        {\includegraphics[width=\tempwidth]{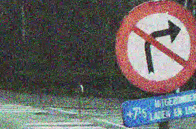}}        
        {\includegraphics[width=\tempwidth]{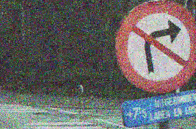}} \\    
    \rowname{Rain}
        {\includegraphics[width=\tempwidth]{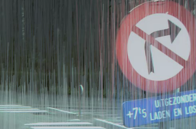}}\vspace{\vblank}
        {\includegraphics[width=\tempwidth]{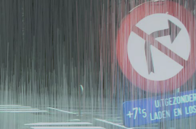}}
        {\includegraphics[width=\tempwidth]{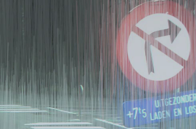}}
        {\includegraphics[width=\tempwidth]{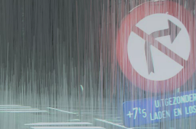}}        
        {\includegraphics[width=\tempwidth]{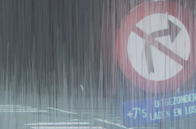}}\\ 
    \rowname{Shadow}
        {\includegraphics[width=\tempwidth]{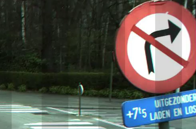}}\vspace{\vblank}
        {\includegraphics[width=\tempwidth]{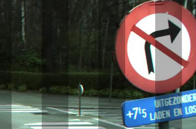}}
        {\includegraphics[width=\tempwidth]{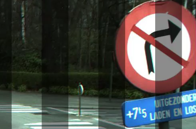}}
        {\includegraphics[width=\tempwidth]{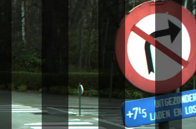}}        
        {\includegraphics[width=\tempwidth]{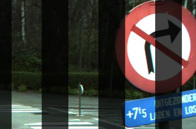}}\\ 
    \rowname{Snow}
        {\includegraphics[width=\tempwidth]{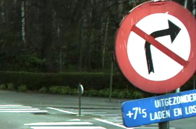}}\vspace{\vblank}
        {\includegraphics[width=\tempwidth]{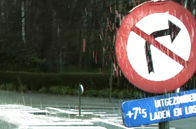}}
        {\includegraphics[width=\tempwidth]{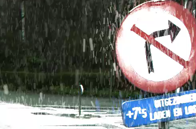}}
        {\includegraphics[width=\tempwidth]{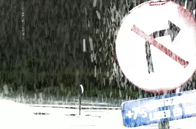}}        
        {\includegraphics[width=\tempwidth]{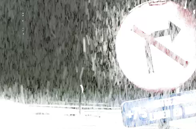}}\\ 
    \rowname{Haze}
        {\includegraphics[width=\tempwidth]{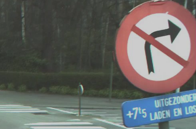}}\vspace{\vblank}
        {\includegraphics[width=\tempwidth]{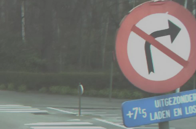}}        
        {\includegraphics[width=\tempwidth]{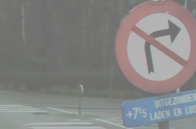}}
        {\includegraphics[width=\tempwidth]{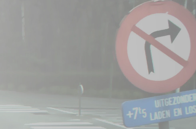}}        
        {\includegraphics[width=\tempwidth]{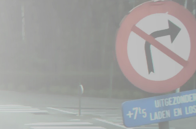}}\\ 

\caption{All challenge types and levels corresponding to a sample frame in the \texttt{CURE-TSD-Real} video sequences.}
\label{fig:challenges}

\end{figure*}
\end{center}
\vspace{-10mm}

\subsection{Challenging Conditions}
\label{subsec_chal_types}
We processed original video sequences with 12 challenge types to obtain challenging video sequences as illustrated in Fig.~\ref{fig:challenges}. Postproduction of challenging conditions scaled up the dataset size from 14,700 images to 896,700 images. We adjusted the level of challenging conditions through visual inspection rather than numerical progression. Challenge-free sequences were considered as level 0 and we added five different levels for each challenge type. Levels were adjusted according to the following rules: \emph{level 1 does not affect the visibility of traffic signs perceptually, level 2 affects the visibility of small and distant traffic signs, level 3 makes the visibility of small and distant traffic signs difficult, level 4 makes the visibility of small and distant traffic signs challenging, and level 5 makes the visibility of  small and distant traffic signs nearly impossible.} Simulated conditions and implementation details are listed as follows:

\begin{itemize}[label=\textcolor{orange}{\FilledSmallSquare},leftmargin=*]

\item $Decolorization$ tests the effect of color acquisition error, which was implemented using \texttt{Black \& White} color correction filter   version 1.0. The filter settings were: \texttt{Reds}$=40$, \texttt{Yellows}$=60$, \texttt{Greens}$=40$, \texttt{Cyans}$=60$, \texttt{Blues}$=20$, and \texttt{Magentas}$=80$. We utilized multiple adjustment layers to compound the effect of the color correction filter and created multiple distinct levels.

\item $Lens/Gaussian~blur$ tests the effect of dynamic scene acquisition. $Lens~blur$ was implemented with the \texttt{Camera Lens Blur} filter version $1.0$ whose radius was set to $2$, $4$, $6$, $8$, and $10$ along with Hexagan \texttt{Iris Shape}. For the $Gaussian~blur$ challenge, \texttt{Gaussian Blur} filter version $3.0$ was used whose \texttt{Bluriness} levels were set to $5, 10, 15, 20,$ and $25$. On contrary to lens~blur, Gaussian~blur is distributed in all directions, which leads to less structured blurred objects.

\item $Codec~error$ tests the effect of encoder/decoder error, which was implemented using \texttt{Time Displacement} filter version $1.6$. \texttt{Max Displacement Time} was set to $0.1, 0.2, 0.3, 0.4,$ and $0.5$.

\item $Darkening~(underexposure)$ tests the effect of underexposure, which was implemented using \texttt{Exposure} filter version $1.0$. The master channel \texttt{Exposure} parameter was set to $-1, -3, -5, -7,$ and $-9$.

\item $Dirty~lens$ tests the effect of occlusion because of dirt over camera lens, which was implemented by overlaying a set of dusty lens images.

\item $Exposure~(overexposure)$ tests the effect of overexpsoure in acquisition, which was implemented with the \texttt{Exposure} filter version $1.0$. The master channel \texttt{Exposure} parameter was set to $1$, $3$, $5$, $7$, and $9$.

\item $Noise$ tests the effect of acquisition noise, which was implemented using the \texttt{Noise} filter version $2.6$. The \texttt{Amount of Noise} parameter was set to $20$, $40$, $60$, $70$, and $71$.

\item $Rain$ tests the effect of occlusion due to rain, which was implemented using the \texttt{Gradient Ramp} generator version $3.2$ with colors \texttt{\# 0F1E2D} and \texttt{\# 5A7492} to create a blueish hue over the video, and \texttt{CC Rainfall} generator from Cycore Effects HD 1.8.2 version $1.1$. The \texttt{Opacity} level was set to $25\%$ with $5$ adjustment layers and the \texttt{Drops} option was set to $10000$, $20000$, $50000$, and $100000$.

\item $Shadow~(occlusion)$ tests the effect of non-uniform lighting due to shadow. Based on the description of Merriam Webster \cite{Webster2019}, shadow refers to \emph{partial darkness or obscurity within a part of space from which rays from a  source of light are cut off by an interposed opaque body}. In this study, darkness/obscurity refers to the vertical patterns and space refers to the traffic sign. In Fig.~\ref{fig:challenges}, we can observe that shadow partially occludes the traffic sign and the levels corresponds to the darkness of the occluded region. The $shadow$ condition was implemented using \texttt{Venetian Blinds} filter version $2.3$. \texttt{Transition Completeness} was $47\%$, \texttt{Direction} was $0x+0.0^{\circ}$, and \texttt{Width} was $142$. Finally, \texttt{Opacity} was set to $15\%, 30\%, 45\%, 60\%$, and $75\%$.

\item $Snow$ tests the effect of occlusion due to snow, which was implemented using the \texttt{Glow} filter version $2.6$ with color \texttt{\# FFFFF} to create a white hue over the video, and \texttt{CC Snowfall} generator from Cycore Effects HD 1.8.2 version $1.1$. \texttt{Glow Threshold} was $55\%$, \texttt{Glow Intensity} was $1.4$, \texttt{Glow Operation} was Screen, and \texttt{Glow Dimension} was Horizontal. \texttt{Drops} option in the \texttt{CC Snowfall} generator was set to $10000, 50000, 100000$, and $140000$ using $9$ adjustment layers.  

\item $Haze$ tests the effect of occlusion due to haze, which was implemented using the \texttt{Ellipse Shape Layer} filter version $1.0$ with radial gradient fill using color \texttt{\# D6D6D6} in the center with $100\%$ opacity and color \texttt{\# 000000} at the edges with $0\%$ opacity. The shape and focal point location of the ellipse was manually controlled to closely follow the furthest point in the video, which created a sense of depth to the scene and emulated the behaviour of haze. In addition to \texttt{Ellipse} filter, \texttt{Smart Blur} version $1.0$, \texttt{Exposure} version $1.0$, and \texttt{Brightness \& Contrast} version $1.0$ were utilized. In the \texttt{Smart Blur} filter, \texttt{Radius} was set to $3$, and \texttt{Threshold} was set to $25$. In the \texttt{Exposure} filter, \texttt{Radius} was $-1$ and \texttt{Gamma Correction} was $1$. In the \texttt{Brightness \& Contrast} filter, \texttt{Brightness} was set to $-34$ and \texttt{Contrast} was set to $-13$. \texttt{Solid Layer} was used with a color code of \texttt{\# CECECE} and opacity was set to $10\%$, $20\%$, $30\%$, $40\%$, and $50\%$.

\end{itemize}

Challenge types were selected and synthesized based on the discussion with the Multimedia Signal Processing Technical Committee and IEEE Signal Processing Society during the VIP Cup 2017 competition process.  In the original competition proposal, we proposed challenging conditions including $blur$, $exposure$, $rain$, and $snow$. Based on the recommendations of the committee members and follow-up discussions, we added the remaining challenging conditions including $decolorization$, $codec~error$, $dirty~lens$, $noise$, $shadow$, and $haze$. All of the challenging conditions other than $snow$ were observed in the prior real-world traffic datasets \cite{Grigorescu2003,Timofte2009,Timofte2014,Belaroussi2010,Larsson2011,Stallkamp2011,Stallkamp2012,Houben2013,Mogelmose2012,Zhu2016,Yang2016,Zhang2017} as summarized in Table \ref{tab_datasets}, which indicates the relevance and significance of these conditions. To simulate challenging conditions,  we utilized the state-of-the-art visual effects and motion graphics software Adobe(c) After Effects, which has been commonly used for realistic image and video processing and synthesis in the literature \cite{Liu2013,Liu2014,Zhuang2017}. We provided the details of challenge generation process including the operators and parameters so that challenging condition synthesis can be reproducible, and researchers can build on top of the initial configurations. Challenging conditions do not have to be used all together and researchers can select the challenging conditions that are relevant and sufficient for their target application. This dataset can be considered as an initial step to assess the robustness of traffic sign detection algorithms under controlled challenging conditions, which can be further enhanced by the research community.  

\vspace{-2mm}

\subsection{Benchmark Algorithms}
In this study, we analyze the average performance of two top performing algorithms in the VIP Cup traffic sign detection challenge \cite{Temel2018_SPM,Temel2019}. Both of the algorithms are based on state-of-the-art convolutional neural networks (CNNs) including U-Net \cite{Ronneberger2015}, ResNet \cite{He2015}, VGG\cite{Simonyan2014}, and GoogLeNet \cite{Szegedy2014}. In both algorithms, localization and recognition of the traffic signs are performed by separate CNNs. Details of the finalist algorithms are summarized as follows: First algorithm includes a VGG-based challenge type detection stage followed by a histogram equalization and a ResNet-based denoising depending on the challenge type. Traffic signs are localized by a U-Net architecture and recognized by a custom CNN architecture. Second algorithm extracts features with a pretrained GoogLeNet architecture whose features at the end of Inception 5B layer are fed into the regression layer to obtain sign locations, which are further classified by a custom CNN architecture.

\begin{table}[htbp!]
\centering
\caption{Detection performance metrics. $\beta$ ($0.5$ or $2.0$) is used to adjust the relative importance of $precision$ and $recall$.}
\begin{tabular}{c|c}
\hline
\textbf{Term} & \textbf{Description/Formulation} \\ \hline
Positive ($P$) & Total number of traffic signs \\ 
True positive ($TP$) & Total number of correct traffic sign detections \\
False positive ($FP$) & Total number of false traffic sign detections \\ 
False negative($FN$) & Number of undetected traffic signs\\
Precision (prec) & $TP/(TP+FP)$ \\
Recall (rec)& $TP/(TP+FN)$ \\
$F_{\beta}$ score & $(1+{\beta}^2)(prec \cdot rec)/({\beta}^2 \cdot prec +rec)$
\\ \hline
 \end{tabular}
 \vspace{-4mm}
\label{tab:metrics}
\end{table}

\section{Traffic Sign Detection under Challenging Conditions}
\label{sec:perf}

\subsection{Performance Metrics}
\label{subsec:perf_metrics}
We calculate precision, recall, $F_{0.5}$, and $F_{2}$ metrics to measure the traffic sign detection performance as described in Table \ref{tab:metrics}. A detection is considered correct if intersection over union (IoU), also known as Jaccard index, is at least 0.5. IoU is obtained by calculating the overlapping area between the ground truth bounding box and the estimated bounding box, and diving the overlaping area by the area of the union between these boxes.

\subsection{Training and Test Sets}
\label{subsec:perf_data}
There are 49 reference video sequences with 300 frames per video as described in Section \ref{subsec:cure_general}. Video sequences were split into training and test sequences by following 7:3 splitting ratio, which led to 34 training video sequences and 15 test video sequences. For each reference video sequence, we generated 60 challenging sequences based on 12 challenge types and 5 challenge levels. For each challenge type, there are 170 (34x5) video sequences (51,000 frames) in the training set and 75 (15x5) video sequences (22,500 frames) in the test set. We set the number of video sequences same for each challenge type and level to eliminate any bias towards a specific challenge type or level. Overall, there are 2,074 video sequences (622,200 frames) in the training set and 915 video sequences (274,500 frames) in the test set.   
\begin{figure}[htbp!]
\begin{minipage}[b]{0.44\linewidth}
  \centering
\includegraphics[width=\linewidth]{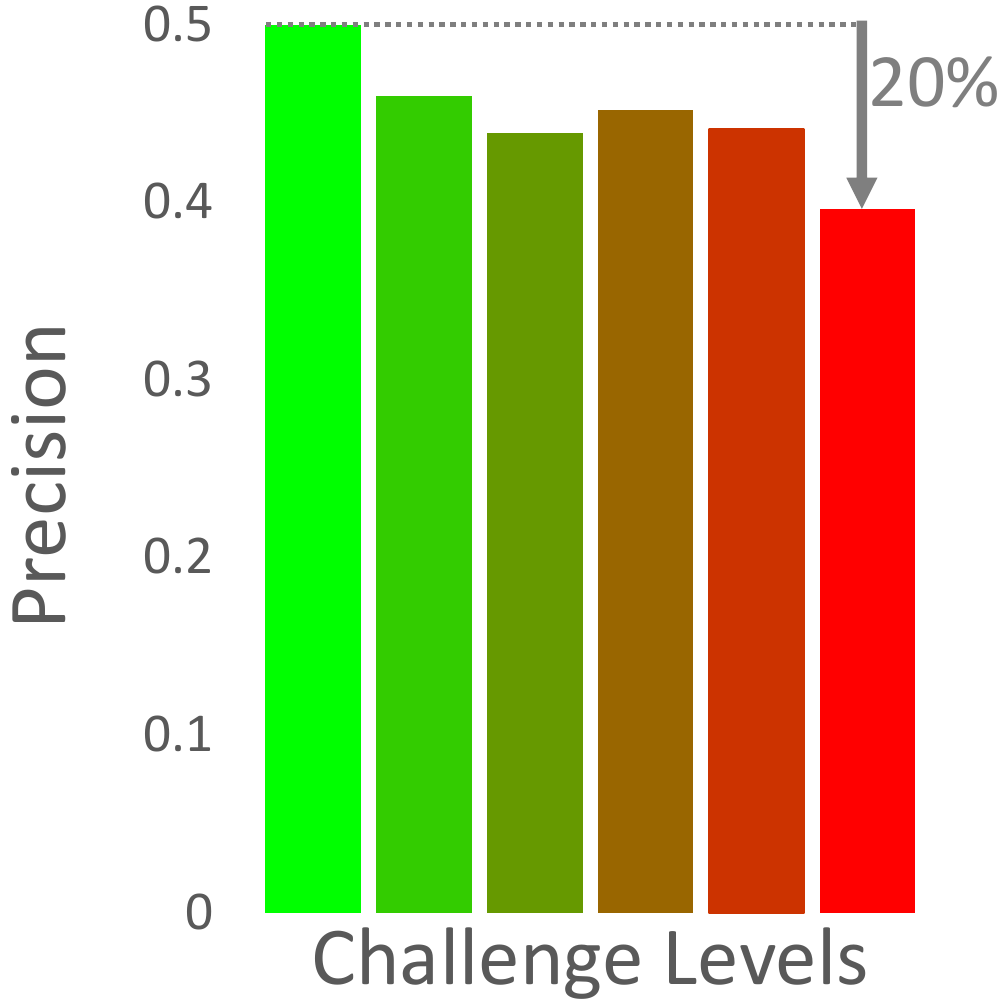}
  \vspace{0.03cm}
  \centerline{\scriptsize{(a) Precision versus challenge levels}}
\end{minipage}
 \vspace{0.05cm}
\begin{minipage}[b]{0.44\linewidth}
  \centering
\includegraphics[width=\linewidth]{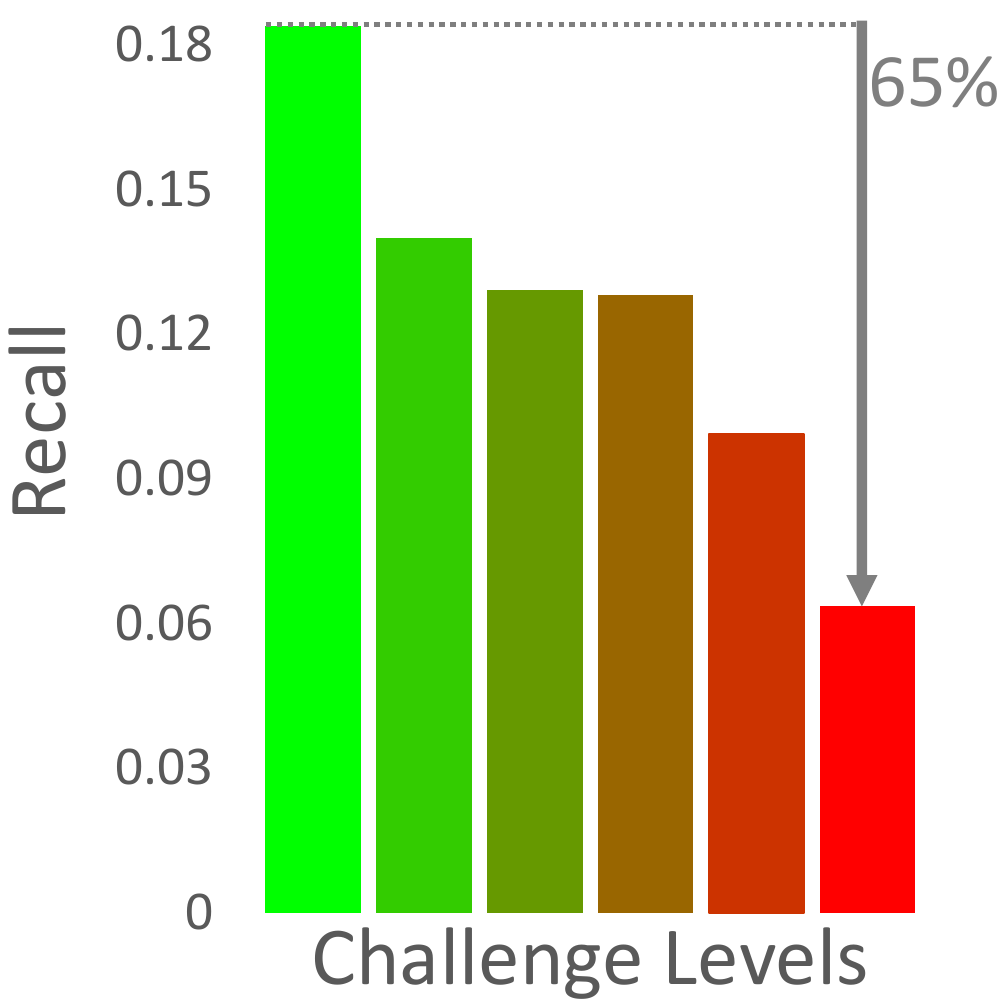}
  \vspace{0.03 cm}
  \centerline{\scriptsize{(b) Recall versus challenge levels} }
\end{minipage}
 \vspace{0.05cm}
\begin{minipage}[b]{0.44\linewidth}
  \centering
\includegraphics[width=\linewidth]{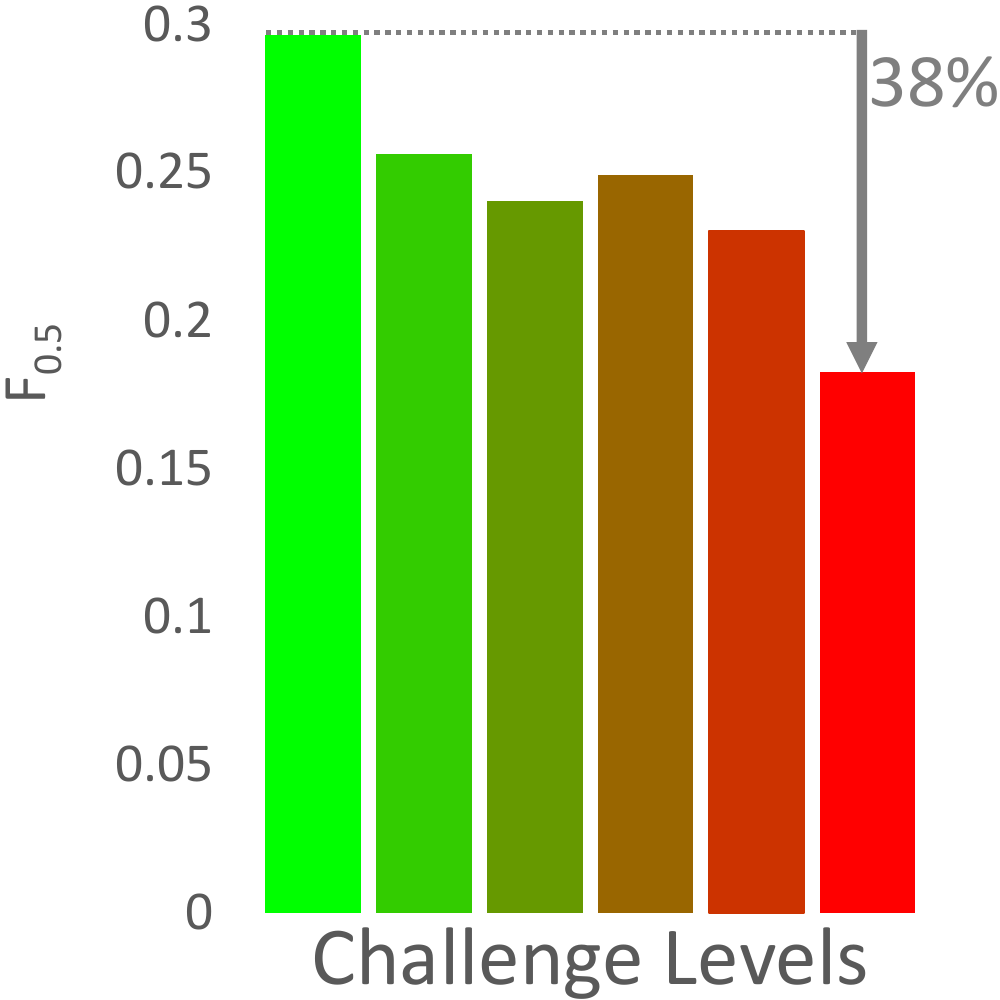}
  \vspace{0.03 cm}
  \centerline{\scriptsize{(c) $F_{0.5}$ versus challenge levels} }
\end{minipage}
 \vspace{0.05cm}
\begin{minipage}[b]{0.44\linewidth}
  \centering
\includegraphics[width=\linewidth]{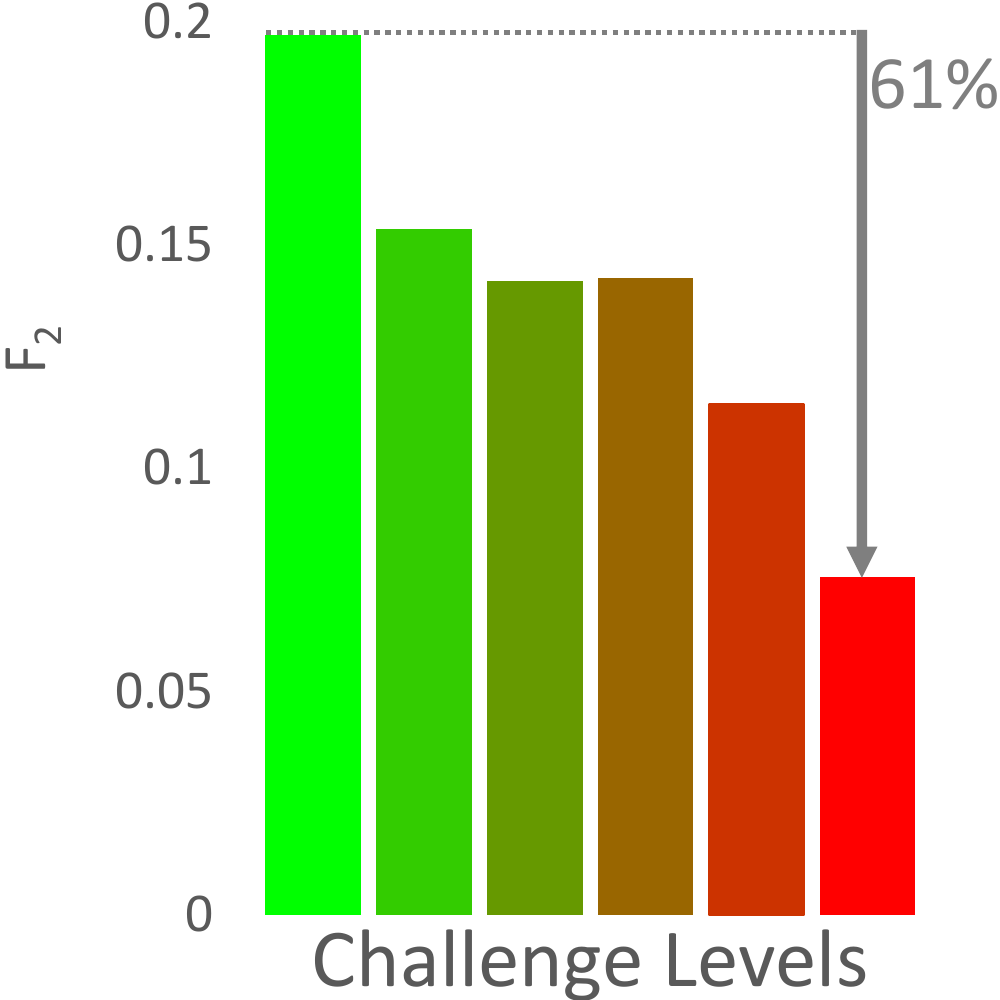}
  \vspace{0.03cm}
  \centerline{\scriptsize{(d) $F_{2}$ versus challenge levels}}
\end{minipage}
 \vspace{0.2cm}
\begin{minipage}[b]{\linewidth}
  \centering
\includegraphics[width=\linewidth]{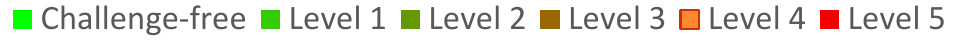}
\end{minipage}
 \caption{Average traffic sign detection performance of top two algorithms with respect to challenge levels over all categories. 
 Performance variations between challenge-free and severe conditions are reported for each metric in percentage.}
\label{fig:overall_levels}
\end{figure}

\begin{table*}[htbp!]
\small
\centering
\caption{Detection performance  of top-I, top-II, and average top-I-II algorithms under challenging conditions for each challenge type and performance metric. First row reports the detection performance over reference video sequences without simulated challenging conditions. For each challenge type, we report the detection performance in the corresponding first row and percentage performance degradation ($\downarrow$) with respect to the challenge-free conditions in the corresponding second row. At the last two rows, we report the average detection performances and performance degradations ($\downarrow$) with bold font over all challenge types.}
\label{tab_variation}
\begin{tabular}{|l|cccc|cccc|cccc|}
\hline
\bf Algorithms &\multicolumn{4}{c}{\textbf{Top-I}} & \multicolumn{4}{c}{\textbf{Top-II}} & \multicolumn{4}{c|}{\textbf{Average Top-I-II}}\\ \hline
\multirow{2}{*}{\textbf{\begin{tabular}[c]{@{}l@{}}Challenge\\Types\end{tabular}}}
& \multirow{2}{*}{\textbf{\begin{tabular}[c]{@{}c@{}}Precision\end{tabular}}}      & \multirow{2}{*}{\textbf{\begin{tabular}[c]{@{}c@{}}Recall\end{tabular}}}         &\multirow{2}{*}{\begin{tabular}[c]{@{}c@{}}$\bm{F_{0.5}}$\end{tabular}}  & \multirow{2}{*}{\begin{tabular}[c]{@{}c@{}}$\bm{F_{2}}$\end{tabular}} & \multirow{2}{*}{\textbf{\begin{tabular}[c]{@{}c@{}}Precision\end{tabular}}}      & \multirow{2}{*}{\textbf{\begin{tabular}[c]{@{}c@{}}Recall\end{tabular}}}         &\multirow{2}{*}{\begin{tabular}[c]{@{}c@{}}$\bm{F_{0.5}}$\end{tabular}}  & \multirow{2}{*}{\begin{tabular}[c]{@{}c@{}}$\bm{F_{2}}$\end{tabular}}
& \multirow{2}{*}{\textbf{\begin{tabular}[c]{@{}c@{}}Precision\end{tabular}}}      & \multirow{2}{*}{\textbf{\begin{tabular}[c]{@{}c@{}}Recall\end{tabular}}}         &\multirow{2}{*}{\begin{tabular}[c]{@{}c@{}}$\bm{F_{0.5}}$\end{tabular}}  & \multirow{2}{*}{\begin{tabular}[c]{@{}c@{}}$\bm{F_{2}}$\end{tabular}}
\\ &&&&&&&&&&&&
\\ \hline

-& 0.35 & 0.29 & 0.34 & 0.30 & 0.65 & 0.07 & 0.25 & 0.09 & 0.50 & 0.18 & 0.30 & 0.20 \\ \hline 
\multirow{2}{*}{\begin{tabular}[c]{@{}c@{}}\bf Decolorization\end{tabular}} & 0.00 & 0.00 & 0.00 & 0.00 & 0.67 & 0.07 & 0.24 & 0.08 & 0.34 & 0.03 & 0.12 & 0.04 \\ 
&$\downarrow$ 100 &$\downarrow$ 100 &$\downarrow$ 100 &$\downarrow$ 100 &4 &$\downarrow$ 10 &$\downarrow$ 6 &$\downarrow$ 10 &$\downarrow$ 32 &$\downarrow$ 82 &$\downarrow$ 60 &$\downarrow$ 79 \\ \hline 
\multirow{2}{*}{\begin{tabular}[c]{@{}c@{}}\bf Lens blur\end{tabular}} & 0.19 & 0.12 & 0.17 & 0.13 & 0.54 & 0.08 & 0.25 & 0.10 & 0.44 & 0.10 & 0.22 & 0.12 \\ 
&$\downarrow$ 45 &$\downarrow$ 59 &$\downarrow$ 49 &$\downarrow$ 57 &$\downarrow$ 17 &11 &1 &10 &$\downarrow$ 12 &$\downarrow$ 45 &$\downarrow$ 24 &$\downarrow$ 41 \\ \hline 
\multirow{2}{*}{\begin{tabular}[c]{@{}c@{}}\bf Codec error\end{tabular}} & 0.04 & 0.02 & 0.03 & 0.02 & 0.17 & 0.01 & 0.04 & 0.01 & 0.11 & 0.01 & 0.03 & 0.01 \\ 
&$\downarrow$ 89 &$\downarrow$ 95 &$\downarrow$ 91 &$\downarrow$ 94 &$\downarrow$ 73 &$\downarrow$ 89 &$\downarrow$ 86 &$\downarrow$ 88 &$\downarrow$ 79 &$\downarrow$ 93 &$\downarrow$ 89 &$\downarrow$ 93 \\ \hline 
\multirow{2}{*}{\begin{tabular}[c]{@{}c@{}}\bf Darkening\end{tabular}} & 0.19 & 0.14 & 0.18 & 0.15 & 0.62 & 0.07 & 0.25 & 0.09 & 0.41 & 0.11 & 0.22 & 0.12 \\ 
&$\downarrow$ 47 &$\downarrow$ 52 &$\downarrow$ 48 &$\downarrow$ 51 &$\downarrow$ 4 &$\downarrow$ 0 &$\downarrow$ 0 &$\downarrow$ 0 &$\downarrow$ 19 &$\downarrow$ 41 &$\downarrow$ 27 &$\downarrow$ 39 \\ \hline 
\multirow{2}{*}{\begin{tabular}[c]{@{}c@{}}\bf Dirty lens\end{tabular}} & 0.10 & 0.04 & 0.07 & 0.04 & 0.60 & 0.07 & 0.24 & 0.08 & 0.36 & 0.05 & 0.16 & 0.06 \\ 
&$\downarrow$ 72 &$\downarrow$ 87 &$\downarrow$ 78 &$\downarrow$ 86 &$\downarrow$ 7 &$\downarrow$ 8 &$\downarrow$ 7 &$\downarrow$ 8 &$\downarrow$ 28 &$\downarrow$ 71 &$\downarrow$ 48 &$\downarrow$ 68 \\ \hline 
\multirow{2}{*}{\begin{tabular}[c]{@{}c@{}}\bf Exposure\end{tabular}} & 0.04 & 0.01 & 0.02 & 0.01 & 0.25 & 0.02 & 0.09 & 0.03 & 0.14 & 0.02 & 0.05 & 0.02 \\ 
&$\downarrow$ 90 &$\downarrow$ 98 &$\downarrow$ 95 &$\downarrow$ 98 &$\downarrow$ 61 &$\downarrow$ 66 &$\downarrow$ 65 &$\downarrow$ 66 &$\downarrow$ 71 &$\downarrow$ 92 &$\downarrow$ 82 &$\downarrow$ 91 \\ \hline 
\multirow{2}{*}{\begin{tabular}[c]{@{}c@{}}\bf Gaussian blur\end{tabular}} & 0.24 & 0.03 & 0.11 & 0.04 & 0.54 & 0.08 & 0.25 & 0.09 & 0.41 & 0.06 & 0.18 & 0.07 \\ 
&$\downarrow$ 33 &$\downarrow$ 89 &$\downarrow$ 68 &$\downarrow$ 87 &$\downarrow$ 17 &6 &$\downarrow$ 2 &5 &$\downarrow$ 18 &$\downarrow$ 70 &$\downarrow$ 40 &$\downarrow$ 66 \\ \hline 
\multirow{2}{*}{\begin{tabular}[c]{@{}c@{}}\bf Noise\end{tabular}} & 0.33 & 0.10 & 0.24 & 0.11 & 0.65 & 0.04 & 0.18 & 0.06 & 0.50 & 0.07 & 0.21 & 0.08 \\ 
&$\downarrow$ 6 &$\downarrow$ 68 &$\downarrow$ 30 &$\downarrow$ 63 &1 &$\downarrow$ 39 &$\downarrow$ 27 &$\downarrow$ 38 &$\downarrow$ 1 &$\downarrow$ 62 &$\downarrow$ 29 &$\downarrow$ 58 \\ \hline 
\multirow{2}{*}{\begin{tabular}[c]{@{}c@{}}\bf Rain\end{tabular}} & 0.00 & 0.00 & 0.00 & 0.00 & 0.45 & 0.05 & 0.17 & 0.06 & 0.22 & 0.02 & 0.09 & 0.03 \\ 
&$\downarrow$ 100 &$\downarrow$ 100 &$\downarrow$ 100 &$\downarrow$ 100 &$\downarrow$ 31 &$\downarrow$ 35 &$\downarrow$ 31 &$\downarrow$ 35 &$\downarrow$ 55 &$\downarrow$ 87 &$\downarrow$ 71 &$\downarrow$ 85 \\ \hline 
\multirow{2}{*}{\begin{tabular}[c]{@{}c@{}}\bf Shadow\end{tabular}} & 0.27 & 0.23 & 0.26 & 0.24 & 0.64 & 0.06 & 0.22 & 0.07 & 0.46 & 0.15 & 0.25 & 0.16 \\ 
&$\downarrow$ 24 &$\downarrow$ 21 &$\downarrow$ 22 &$\downarrow$ 21 &$\downarrow$ 1 &$\downarrow$ 17 &$\downarrow$ 13 &$\downarrow$ 17 &$\downarrow$ 8 &$\downarrow$ 21 &$\downarrow$ 16 &$\downarrow$ 20 \\ \hline 
\multirow{2}{*}{\begin{tabular}[c]{@{}c@{}}\bf Snow\end{tabular}} & 0.28 & 0.04 & 0.13 & 0.05 & 0.60 & 0.06 & 0.22 & 0.08 & 0.44 & 0.05 & 0.17 & 0.06 \\ 
&$\downarrow$ 20 &$\downarrow$ 87 &$\downarrow$ 63 &$\downarrow$ 85 &$\downarrow$ 6 &$\downarrow$ 14 &$\downarrow$ 12 &$\downarrow$ 14 &$\downarrow$ 11 &$\downarrow$ 72 &$\downarrow$ 41 &$\downarrow$ 68 \\ \hline 
\multirow{2}{*}{\begin{tabular}[c]{@{}c@{}}\bf Haze\end{tabular}} & 0.22 & 0.00 & 0.00 & 0.00 & 0.64 & 0.07 & 0.25 & 0.09 & 0.44 & 0.04 & 0.13 & 0.05 \\ 
&$\downarrow$ 37 &$\downarrow$ 100 &$\downarrow$ 99 &$\downarrow$ 100 &$\downarrow$ 1 &$\downarrow$ 2 &$\downarrow$ 0 &$\downarrow$ 1 &$\downarrow$ 11 &$\downarrow$ 79 &$\downarrow$ 56 &$\downarrow$ 77 \\ \hline 
\multirow{2}{*}{\begin{tabular}[c]{@{}c@{}}\bf Average\end{tabular}} &\bf 0.16 &\bf 0.06 &\bf 0.10 &\bf 0.07 &\bf 0.53 &\bf 0.06 &\bf 0.20 &\bf 0.07 &\bf 0.36 &\bf 0.06 &\bf 0.15 &\bf 0.07 \\ 
&\bf $\downarrow$ 55 &\bf $\downarrow$ 80 &\bf $\downarrow$ 70 &\bf $\downarrow$ 78 &\bf $\downarrow$ 18 &\bf $\downarrow$ 22 &\bf $\downarrow$ 21 &\bf $\downarrow$ 22 &\bf$\downarrow$ 29 &\bf $\downarrow$ 68 &\bf $\downarrow$ 48 &\bf $\downarrow$ 65 \\ \hline

\end{tabular}
\end{table*}

\subsection{Performance Variation under Challenging Conditions}
\label{subsec:perf_overall}
As reference performance, we calculate the average detection performance of benchmarked algorithms for challenge-free sequences. Then, we calculate the detection performance under varying challenging conditions and levels. Average detection performance under different challenge levels are reported in Fig.~\ref{fig:overall_levels}. Y-axis corresponds to detection performance and x-axis corresponds to challenge levels. In addition to reporting average detection performance for each challenge level, we also report the percentage performance degradation under severe challenging conditions (level 5). Based on the results in Fig.~\ref{fig:overall_levels}, detection performance significantly decreases with respect to reference challenge-free conditions. Specifically, detection performance degrades by $20\%$ in precision, $65\%$ in recall, $38\%$ in $F_{0.5}$ score and $61\%$ in $F_{2}$ score.

To understand the effect of challenging condition types over traffic sign detection, we report the performance of top-I team, top-II team, and their average (top-I-II) in Table~\ref{tab_variation}. The detection performance of top-I team degrades by $55\%$ in precision, $80\%$ in recall, $70\%$ in $F_{0.5}$  score, and $78\%$ in $F_{2}$ score. For top-II team, the performance degradation is $18\%$ in precision, $22\%$ in recall, $21\%$ in $F_{0.5}$  score, and $22\%$ in $F_{2}$ score. We can observe that overall performance degradation for each team vary between $18\%$ and $80\%$ in which the variation is higher for top-I team. When team results are averaged (Top-I-II), overall performance degrades by $29\%$ in precision, $68\%$ in recall, $48\%$ in $F_{0.5}$  score, and $65\%$ in $F_{2}$ score. Precision under $noise$ challenge is the only category in which performance remains almost same as reported in top-I-II results. In other challenge categories, precision degradation varies between $8\%$ and $79\%$. Performance degradation in recall is more significant than precision in all challenge categories, which varies between $21\%$ and $93\%$.  Performance variation in $F$ score is in between $16\%$ and $93\%$.

Even though almost all the simulated challenging conditions degrade detection performance, there are few exceptions for top-II algorithm in $decolorization$ ($+4\%$~precision), $noise$ ($+1\%$~precision) $lens~blur$ ($+11\%$~recall, $+10\%~F_{2}$, and $+1\%~F_{0.5}$  ), and $Gaussian~blur$ ($+6\%$~recall and $+5\%~F_{2}$). The effect of $decolorization$ significantly depends on the benchmark algorithm and the increase in precision under $noise$ for top-II algorithm does not exceed $1\%$. In the $decolorization$ category, detection performance of top-I algorithm degrades $100\%$ in term of all metrics whereas performance variation of top-II algorithm does not exceed $10\%$. When traffic signs under severe challenging conditions are compared in Fig.~\ref{fig:challenges}, we can observe that structural information including high frequency components mostly remain intact in $decolorization$ category whereas chroma information is distorted. Therefore, we can conclude that top-I algorithm significantly relies on color information whereas top-II does not rely as much. Challenging conditions based on blur filter out high frequency components that can be used to recognize and localize traffic signs. But, at the same time, filtering out high frequency components can eliminate certain false detections, which can explain the minor performance enhancement in the aforementioned exceptional categories.

\begin{figure}[htbp!]
    \centering
    \includegraphics[width=\linewidth]{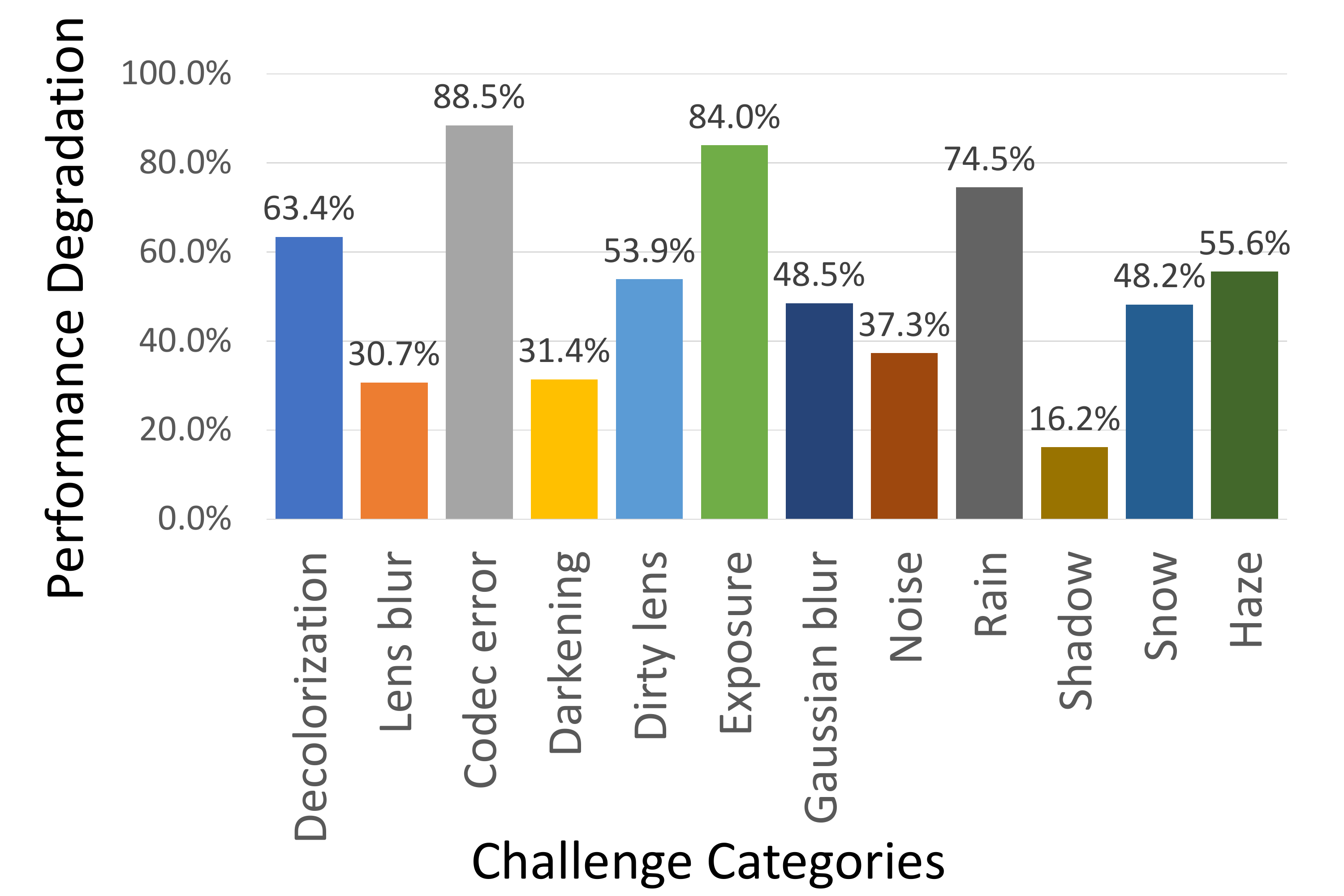}
    \caption{Average detection performance degradation under challenging conditions for each challenge type.}
    \label{fig:perf_deg}
    \vspace{-4 mm}
\end{figure}

We report the average performance degradation over all the performance metrics and algorithms to understand the overall effect of challenging conditions types in Fig.~\ref{fig:perf_deg}. $Codec~error$ and $exposure$ result in the highest performance degradation in the challenge categories. As observed in sixth row from top in Fig.~\ref{fig:challenges}, $exposure$ condition can significantly saturate descriptive regions of traffic signs that are critical for recognition, which results in a high performance degradation. In case of $codec~error$, we observe visual artifacts that corrupt the structural characteristics of the sign. In addition to the structural artifacts, $codec~error$ challenge can relocate a significant portion of the traffic sign to a new location as shown in the third row from top in Fig.~\ref{fig:challenges}, which would not satisfy the required overlap between the ground truth location and the detected location even though the sign can be recognized accurately. Benchmark algorithms can be vulnerable to challenging weather conditions including $rain$, $haze$, and $snow$, which lead to a performance degradation between $48.2\%$ and $74.5\%$.  In real-world scenarios, outer surface of the camera lens or the window surface in front of the camera can get dirty because of the weather and road conditions, which can affect the visibility of traffic signs due to occlusion. Based on the $dirty~lens$ experiments, occlusions can reduce the overall traffic sign detection performance by half. $Blur$, $darkening$, and $noise$ challenges degrade the detection performance between $30.7$ and $48.5\%$ whereas $shadow$ results in the least performance degradation by $16.2\%$. Because of the difficulty of realistic shadow generation, we used a simple periodic pattern to simulate local $shadow$ effect, which can be considered as a partial $darkening$ effect. We can observe that the performance degradation in the $darkening$ category is almost double the degradation in the $shadow$ category, which is proportional to the ratio of the darkened regions when degraded images are compared in both categories as observed in fourth row and tenth row from top in  Fig.~\ref{fig:challenges}. When traffic sign images with high level $darkening$ conditions are observed in Fig.~\ref{fig:challenges}, images may appear almost entirely dark and can be considered as the most challenging condition perceptually. However, the perceptual level of darkness depends on the display settings and if images are observed under different display configurations, it can be possible to observe descriptive parts of the traffic sign even under high level $darkening$ conditions.

\subsection{Spectral Analysis of Challenging Conditions} 
\label{subsec:perf_spectrum}

\begin{figure}[htbp!]
\begin{minipage}[b]{0.32\linewidth}
  \centering
\includegraphics[width=\linewidth, trim= 8mm 10mm 18mm 8mm]{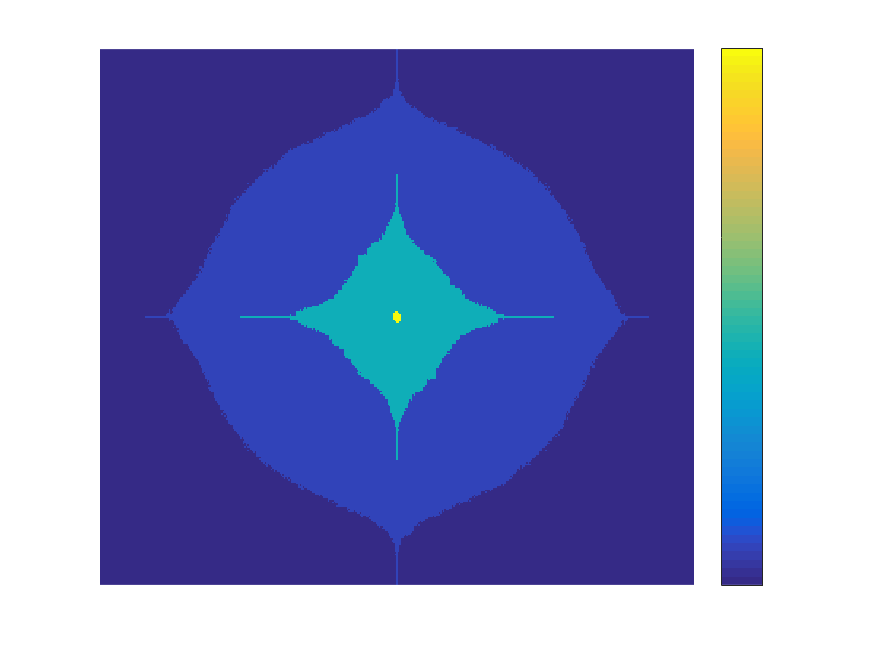}
  \vspace{0.03cm}
  \centerline{\scriptsize{(a) Decolorization}}
\end{minipage}
 \vspace{0.05cm}
\begin{minipage}[b]{0.32\linewidth}
  \centering
\includegraphics[width=\linewidth, trim= 8mm 10mm 18mm 8mm]{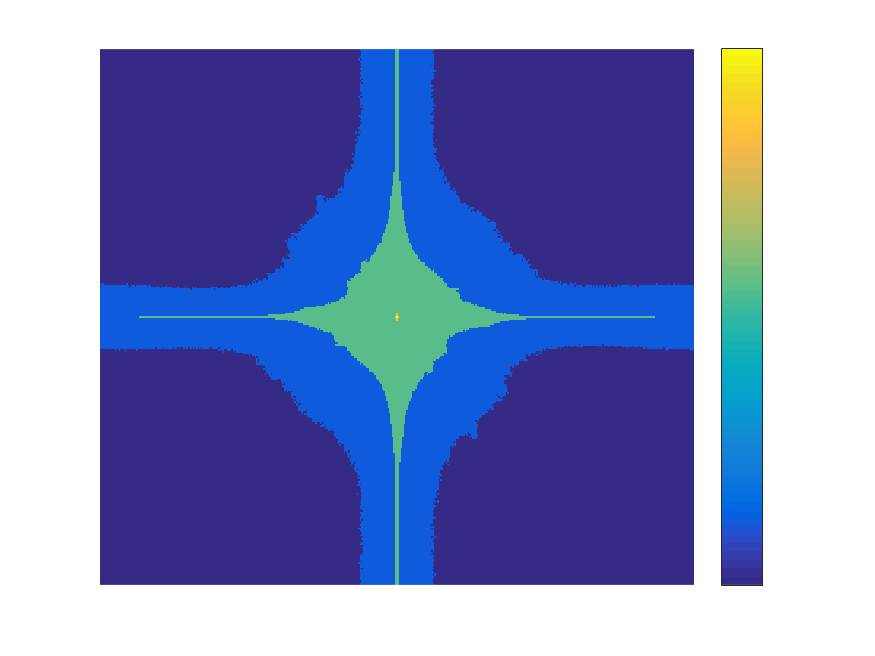}
  \vspace{0.03 cm}
  \centerline{\scriptsize{(b) Lens blur} }
\end{minipage}
 \vspace{0.05cm}
\begin{minipage}[b]{0.32\linewidth}
  \centering
\includegraphics[width=\linewidth, trim= 8mm 10mm 18mm 8mm]{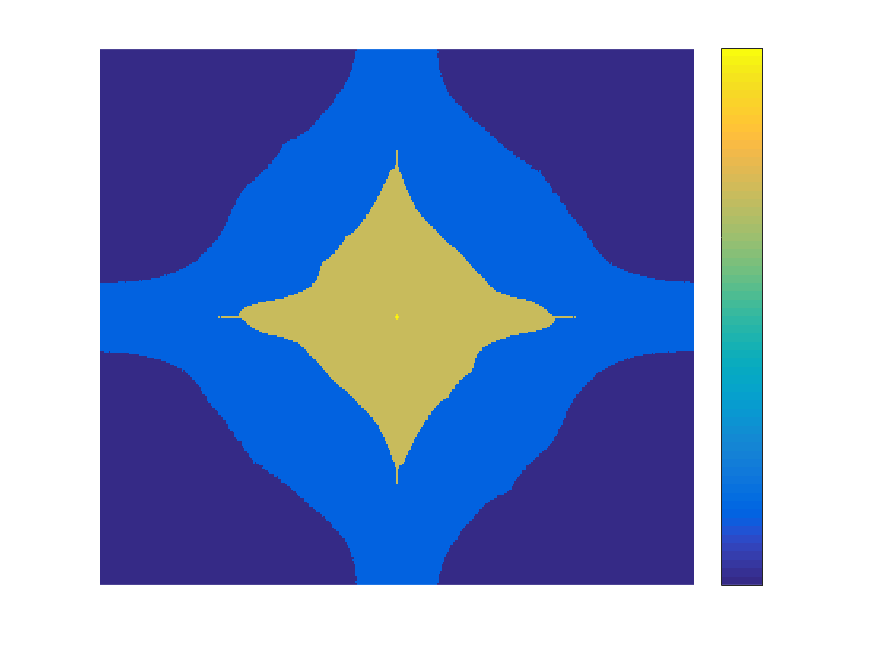}
  \vspace{0.03 cm}
  \centerline{\scriptsize{(c) Codec error} }
\end{minipage}
 \vspace{0.05cm}
\begin{minipage}[b]{0.32\linewidth}
  \centering
\includegraphics[width=\linewidth, trim= 8mm 10mm 18mm 4mm]{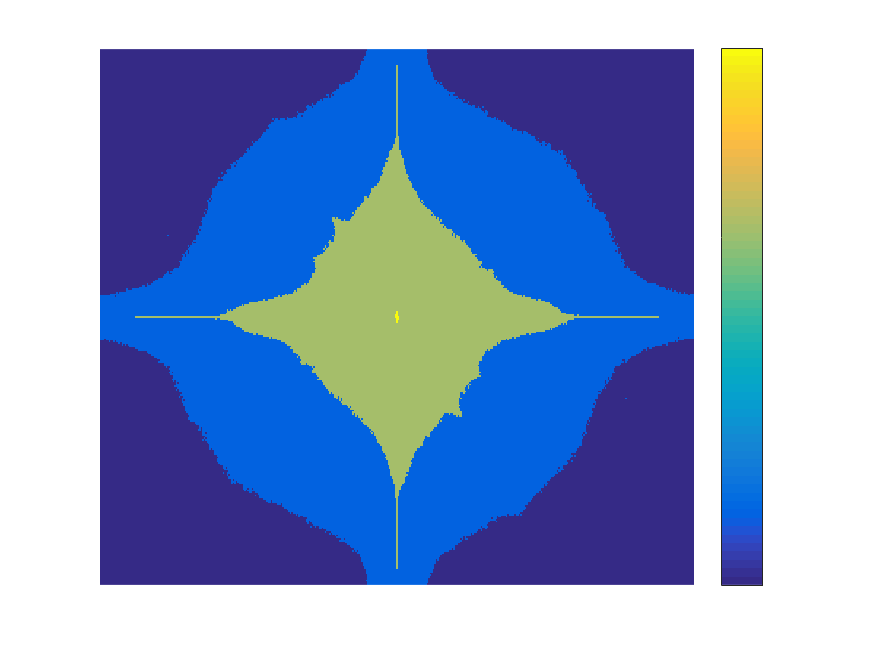}
  \vspace{0.03cm}
  \centerline{\scriptsize{(d) Darkening}}
\end{minipage}
 \vspace{0.05cm}
\begin{minipage}[b]{0.32\linewidth}
  \centering
\includegraphics[width=\linewidth, trim= 8mm 10mm 18mm 4mm]{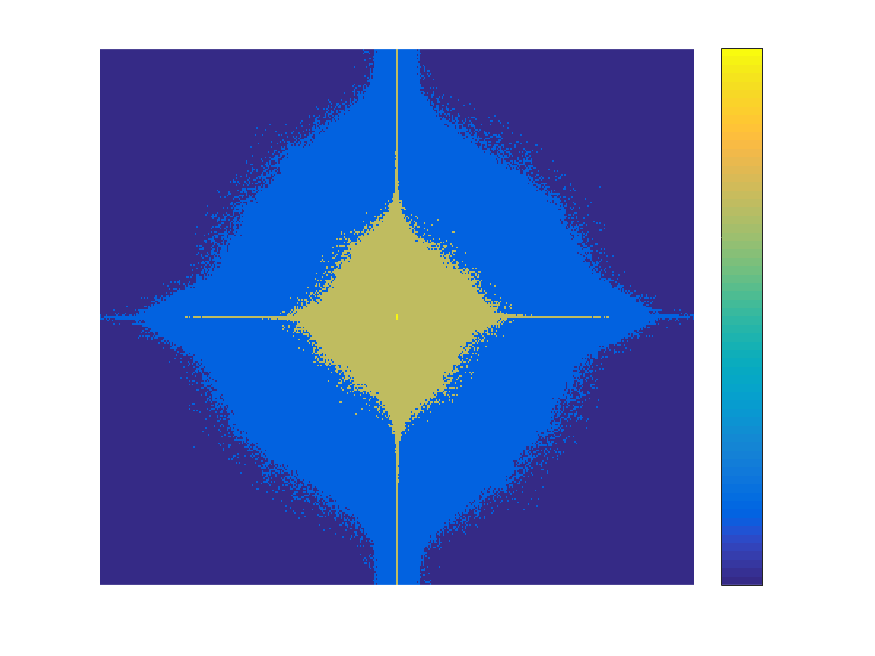}
  \vspace{0.03cm}
  \centerline{\scriptsize{(e) Dirty lens}}
\end{minipage}
 \vspace{0.05cm}
\begin{minipage}[b]{0.32\linewidth}
  \centering
\includegraphics[width=\linewidth, trim= 8mm 10mm 18mm 4mm]{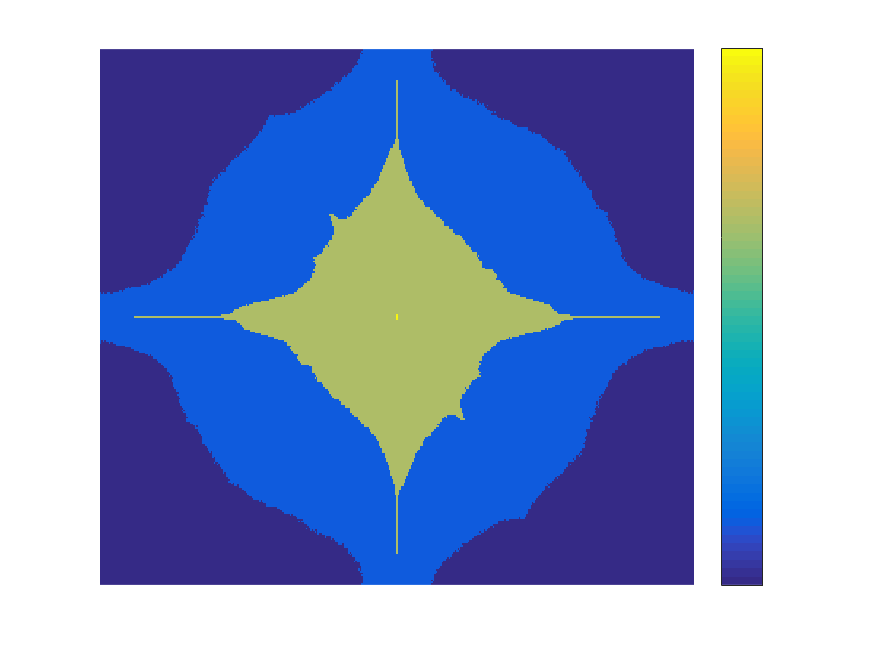}
  \vspace{0.03 cm}
  \centerline{\scriptsize{(f) Exposure} }
\end{minipage}
 \vspace{0.05cm}
\begin{minipage}[b]{0.32\linewidth}
  \centering
\includegraphics[width=\linewidth, trim= 8mm 10mm 18mm 8mm]{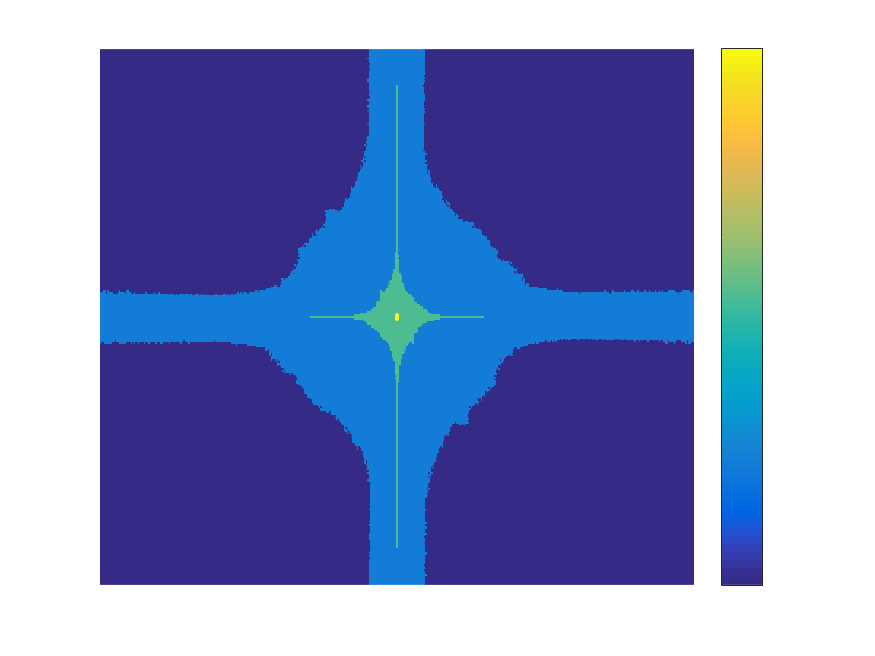}
  \vspace{0.03 cm}
  \centerline{\scriptsize{(g) Gaussian blur} }
\end{minipage}
 \vspace{0.05cm}
\begin{minipage}[b]{0.32\linewidth}
  \centering
\includegraphics[width=\linewidth, trim= 8mm 10mm 18mm 8mm]{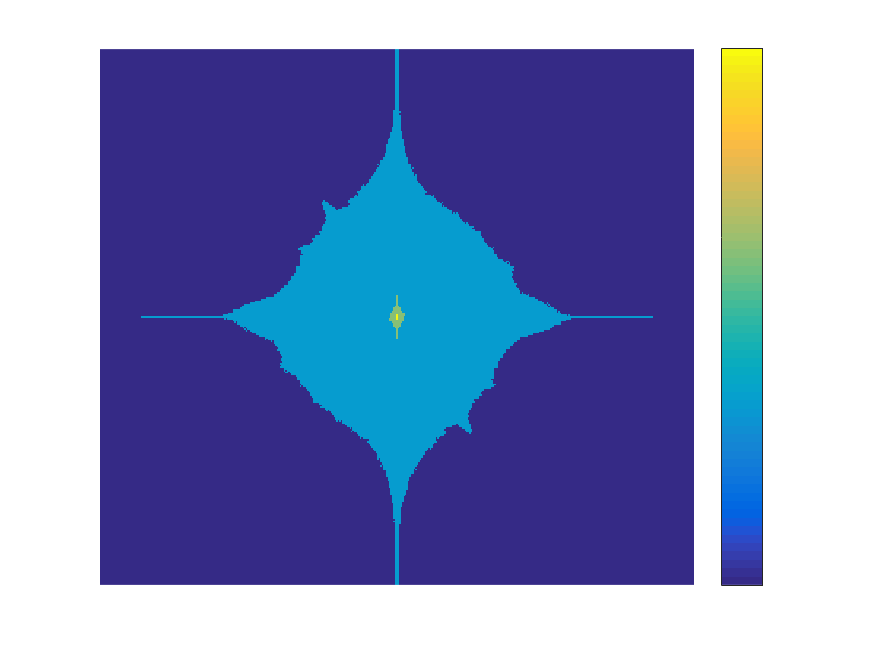}
  \vspace{0.03cm}
  \centerline{\scriptsize{(h) Noise}}
\end{minipage}
 \vspace{0.05cm}
\begin{minipage}[b]{0.32\linewidth}
  \centering
\includegraphics[width=\linewidth, trim= 8mm 10mm 18mm 8mm]{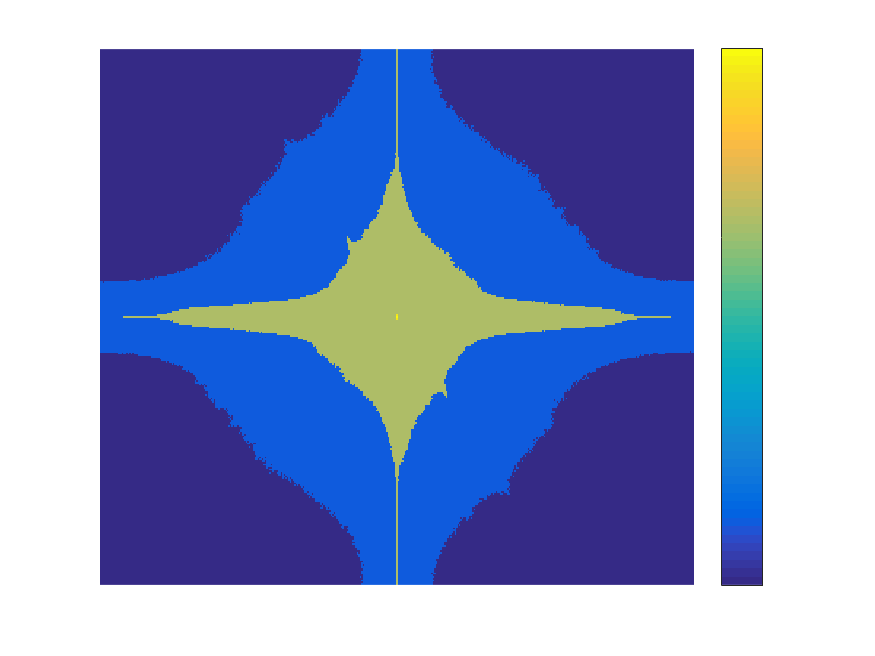}
  \vspace{0.03cm}
  \centerline{\scriptsize{(i) Rain}}
\end{minipage}
 \vspace{0.05cm}
\begin{minipage}[b]{0.32\linewidth}
  \centering
\includegraphics[width=\linewidth, trim= 8mm 10mm 18mm 4mm]{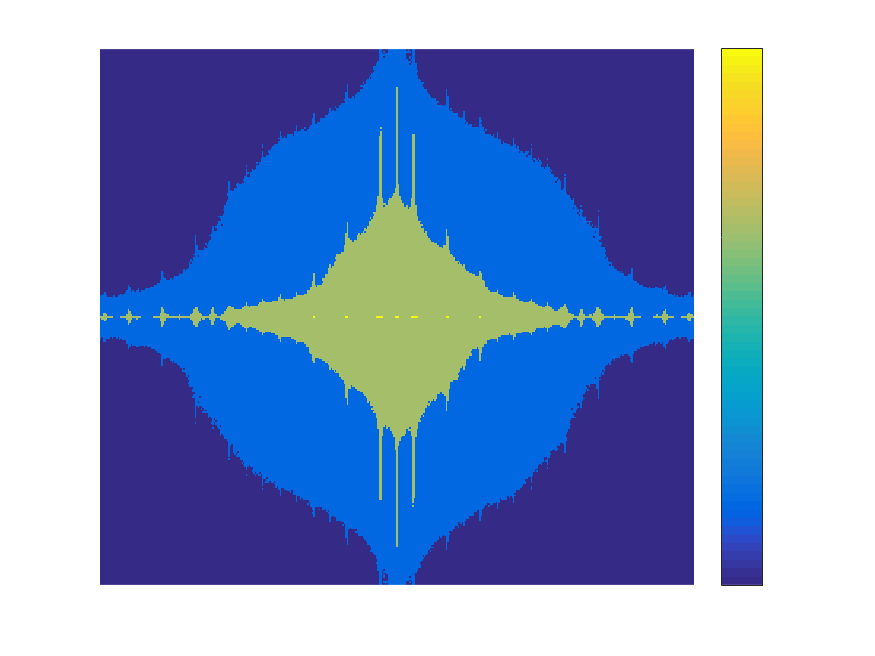}
  \vspace{0.03 cm}
  \centerline{\scriptsize{(j) Shadow} }
\end{minipage}
 \vspace{0.05cm}
\begin{minipage}[b]{0.32\linewidth}
  \centering
\includegraphics[width=\linewidth, trim= 8mm 10mm 18mm 4mm]{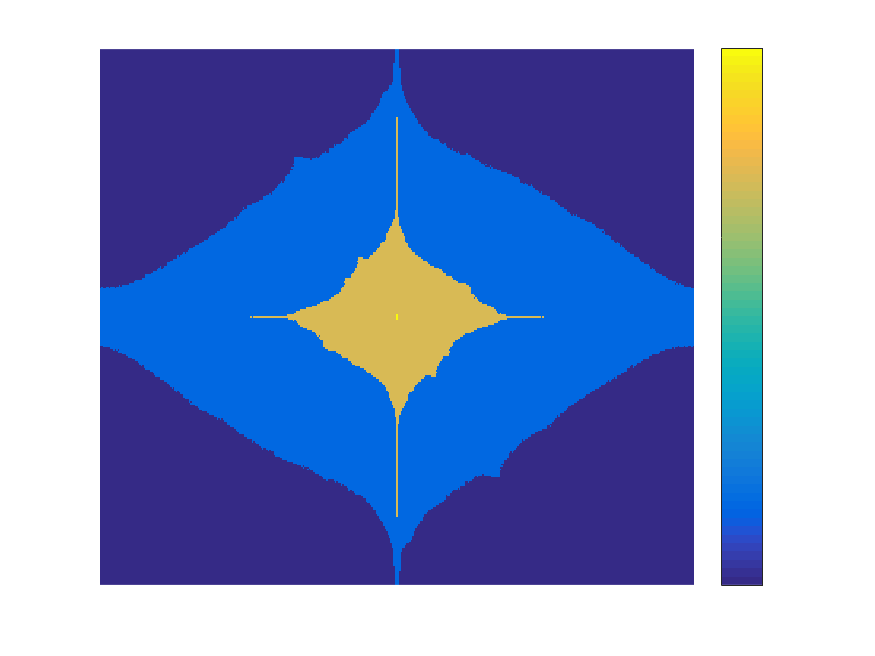}
  \vspace{0.03 cm}
  \centerline{\scriptsize{(k) Snow} }
\end{minipage}
 \vspace{0.05cm}
\begin{minipage}[b]{0.32\linewidth}
  \centering
\includegraphics[width=\linewidth, trim= 8mm 10mm 18mm 4mm]{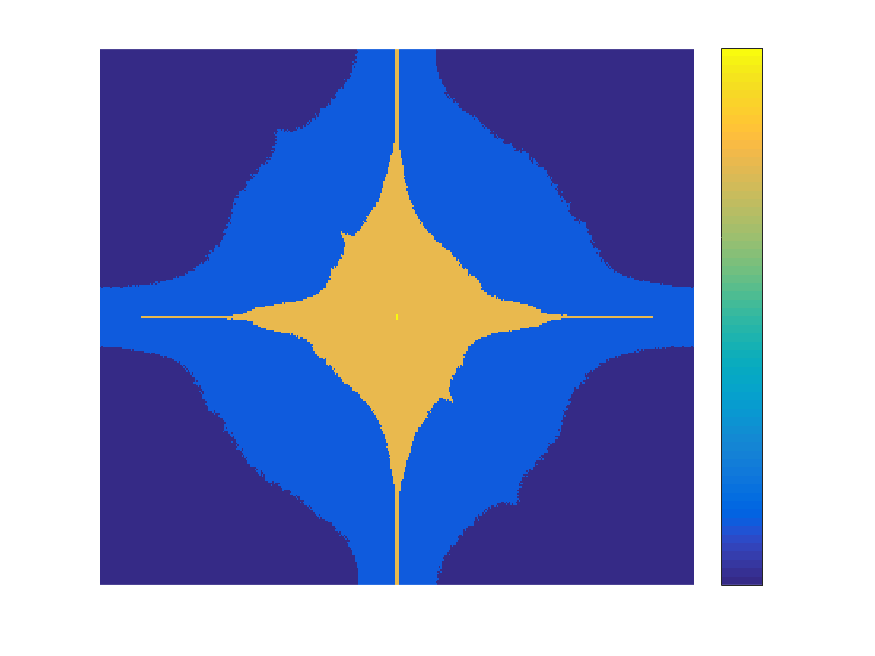}
  \vspace{0.03 cm}
  \centerline{\scriptsize{(l) Haze}}
\end{minipage}
 \caption{Average magnitude spectrum maps of video sequences corresponding to different challenge types.}
\label{fig:spectrum}
\vspace{-4 mm}
\end{figure}

 In this section, we investigate the effect of challenging conditions over the spectral characteristics of video sequences. In the \texttt{CURE-TSD-Real} dataset, there are $49$ challenge-free sequences. Corresponding to each challenge-free reference video, there are $60$ video sequences with distinct challenge conditions  ($12$ types $\times$ $5$ levels). For each challenge-free reference video, we obtain the pixel-wise and frame-wise difference between the reference video and challenging video, which results in the residual video. In total, we obtain $2,940$ ($49$ videos $\times$ $60$ challenge configurations) residual videos, which correspond to $245$ residual videos per challenge type. For each residual video, we calculate the power spectrum per frame to obtain a power spectrum sequence. We calculate the power spectrum of a residual frame as
\begin{equation}
  Log \left\{\left| {\cal F} \left\{\left| R \right| \right\}  \right| \right\}   ,
\label{eq:spectrum}
\end{equation}
where $R$ is the residual frame, $|$  $|$ is the absolute value, ${\cal F}$ is the $2$-$D$ discrete Fourier transform, and $Log$ is the logarithm. We calculate the pixel-wise average of $245$ power spectrum sequences for each challenge type and quantize them to obtain the average magnitude spectrum maps in Fig.~\ref{fig:spectrum}.

In the average magnitude spectrum maps, central region corresponds to low frequency components and corners represent high frequency components. Color coding is based on the magnitude of the frequency components. Color of the spectrum elements varies from dark blue to yellow as their magnitude increases. We can observe that challenging conditions lead to characteristic spectral shapes that can be used to analyze the effect of these conditions. Even though $exposure$ and $darkening$ correspond to perceptually very distinct images as observed in Fig.~\ref{fig:challenges}, their spectral representations correspond to an almost identical pattern in Fig.~\ref{fig:spectrum}(d) and Fig.~\ref{fig:spectrum}(f). Spectral representation of $exposure$ challenge indicates that high-frequency components remain similar to the challenge-free sequences whereas low-frequency components get significantly distorted. $Exposure$ and $haze$ result in non-uniform deformations that affect the visibility of certain regions in an image, which lead to similar spectral representations. In the $blur$ challenge, lens blur and Gaussian blur lead to a similar pattern in which boundaries of horizontal and vertical regions correspond to cutoff frequencies as observed in Fig.~\ref{fig:spectrum}(b) and Fig.~\ref{fig:spectrum}(g). Challenging conditions result in dominant vertical patterns in the $rain$ and the $shadow$ challenges as observed in Fig.~\ref{fig:challenges}, which correspond to a more predominant horizontal pattern in spectral representations as shown in Fig.~\ref{fig:spectrum}(i) and Fig.~\ref{fig:spectrum}(j). Moreover, we observe discrete peeks in the spectral representations of the $shadow$ challenge in Fig.~\ref{fig:spectrum}(j) because of the periodic shadow patterns. In the $rain$ challenge, falling particles are the main occluding factor whereas in the $snow$ challenge, piled up snow significantly occludes certain regions, which limits the highest spectral components to a more central region as observed in Fig.~\ref{fig:spectrum}(k). $Noise$ and $decolorization$ challenge lead to a peak at DC along with minor low frequency degradations as shown in Fig.~\ref{fig:spectrum}(a) and Fig.~\ref{fig:spectrum}(h). In the $codec~error$ challenge, we observe local shifts of certain regions in the images as shown in Fig.~\ref{fig:challenges}, which leads to an almost symmetric spectral representation in Fig.\ref{fig:spectrum}(c) without the sharp horizontal and vertical lines. On contrary to all other challenges, $dirty~lens$ challenge results in deformation over the images that varies in terms of shape and size, which leads to a granular structure in the spectral representation as shown in Fig.~\ref{fig:spectrum}(e). 


\newcommand{\columnname}[1]
{\makebox[\tempwidth][c]{\textbf{#1}}}

\newcommand{\rowname}[1]
{\begin{minipage}[b]{0.5cm}
    \centering\rotatebox{90}{\makebox[\tempheight][c]{\textbf{#1}}}
\end{minipage}}

\begin{figure}[!htbp]

    \setlength{\tempwidth}{.27\linewidth}
    \setlength{\vblank}{0mm}
    \settoheight{\tempheight}{\includegraphics[width=\tempwidth]{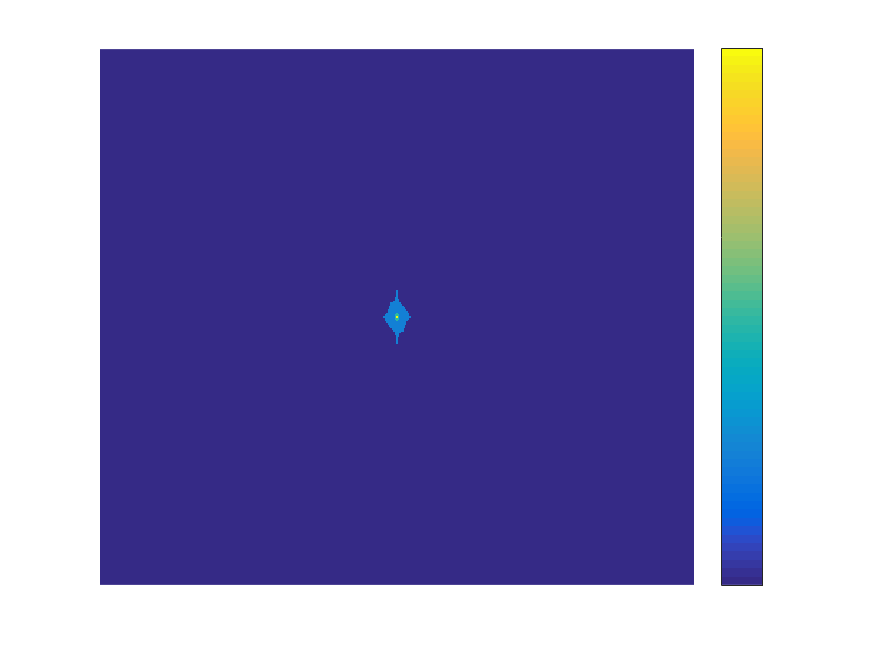}}
    \begin{minipage}[b]{0.2cm}\centering{\makebox[0.2cm][c]{}}\end{minipage}
        \begin{minipage}[c][0.3cm][t]{1.15\tempwidth}\columnname{Minor condition}\end{minipage}\hfill
        \begin{minipage}[c][0.3cm][t]{1.15\tempwidth}\columnname{Medium condition}\end{minipage}\hfill
        \begin{minipage}[c][0.3cm][t]{1.15\tempwidth}\columnname{Major condition}\end{minipage}
    \rowname{Decolor.}
        {\includegraphics[width=\tempwidth]{./Figs/mapsLevels/type1_level1.png}}\vspace{\vblank}\hfill
        {\includegraphics[width=\tempwidth]{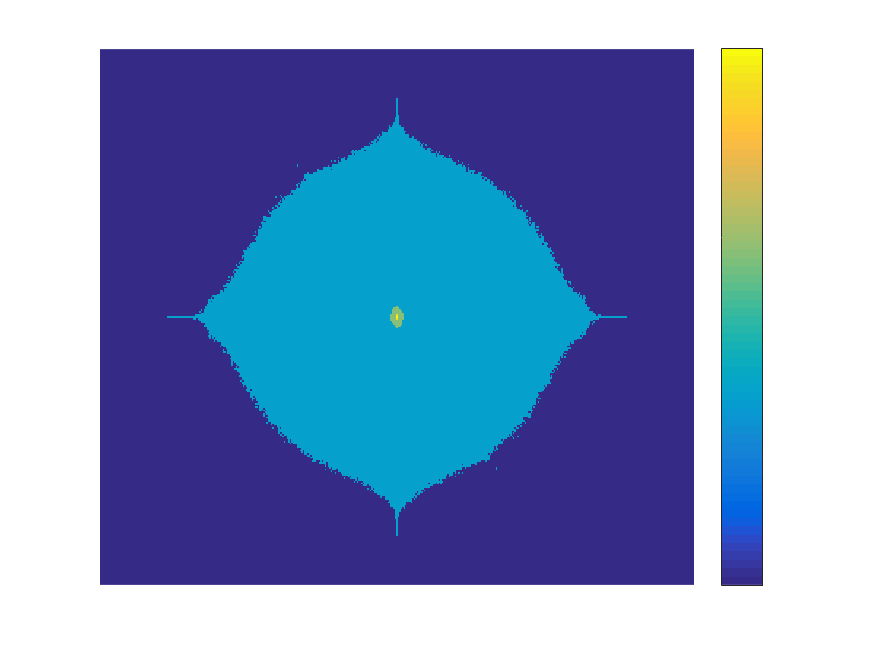}}\hfill
        {\includegraphics[width=\tempwidth]{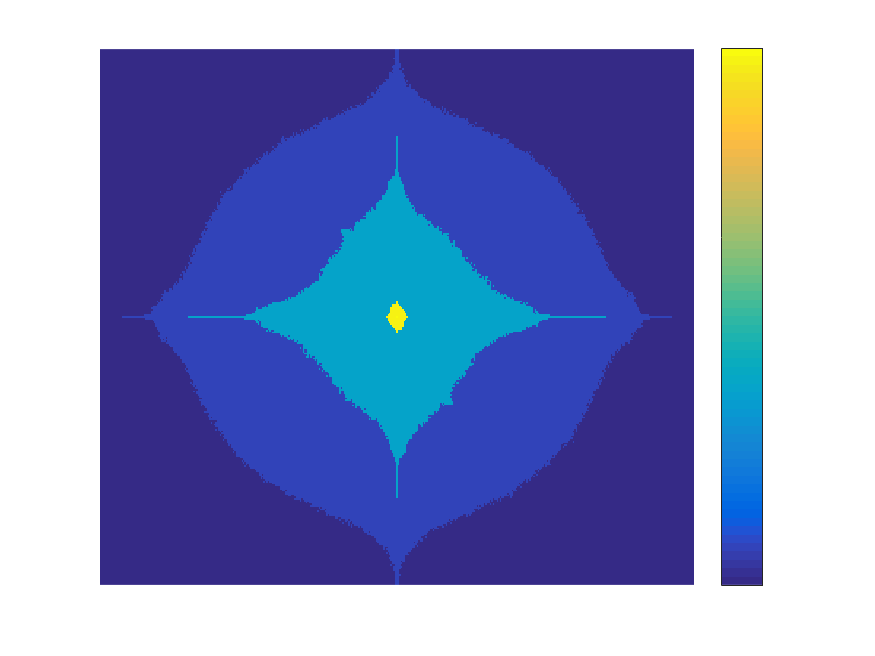}}\\
    \rowname{Lens blur}
        {\includegraphics[width=\tempwidth]{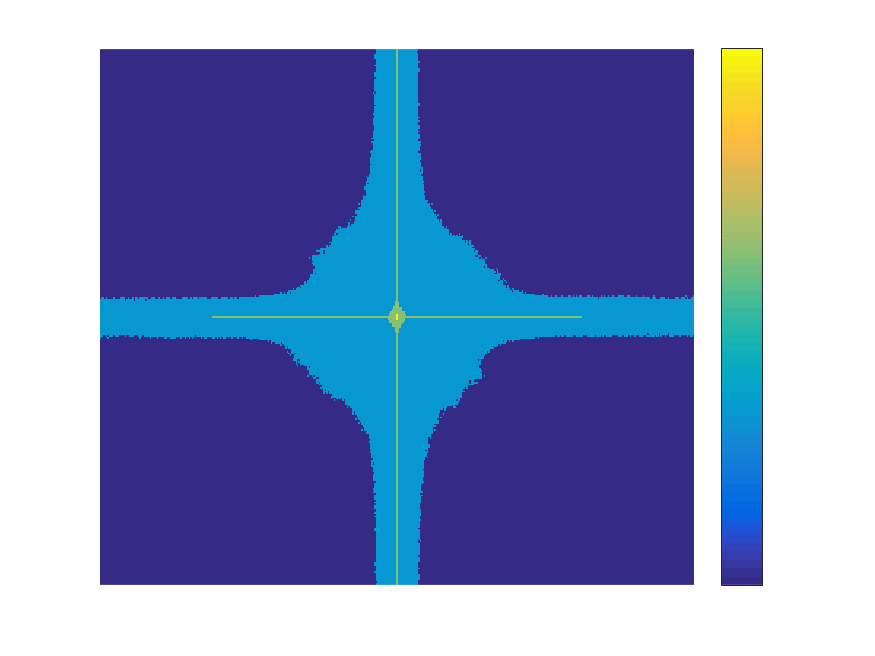}}\vspace{\vblank}\hfill
        {\includegraphics[width=\tempwidth]{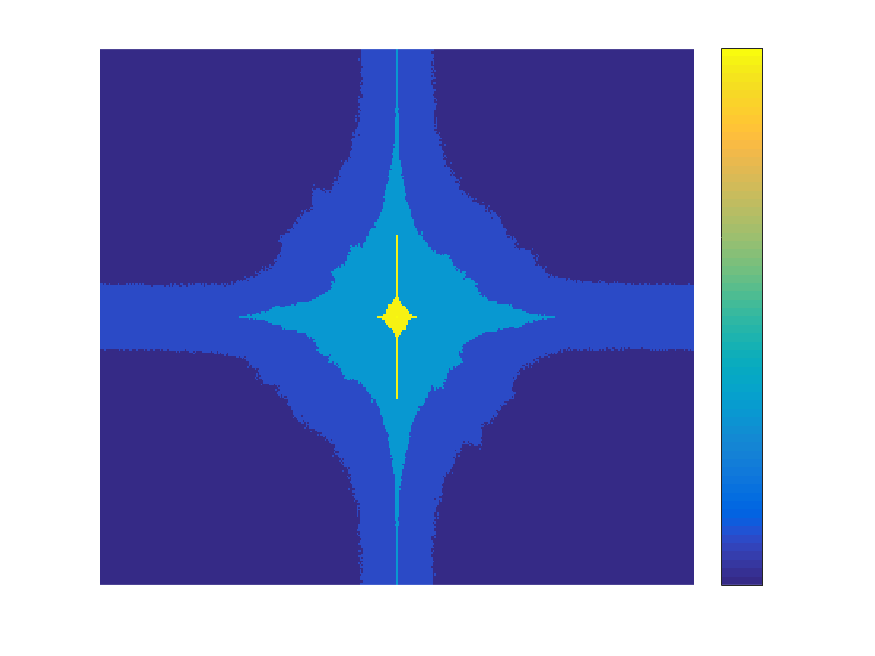}}\hfill
        {\includegraphics[width=\tempwidth]{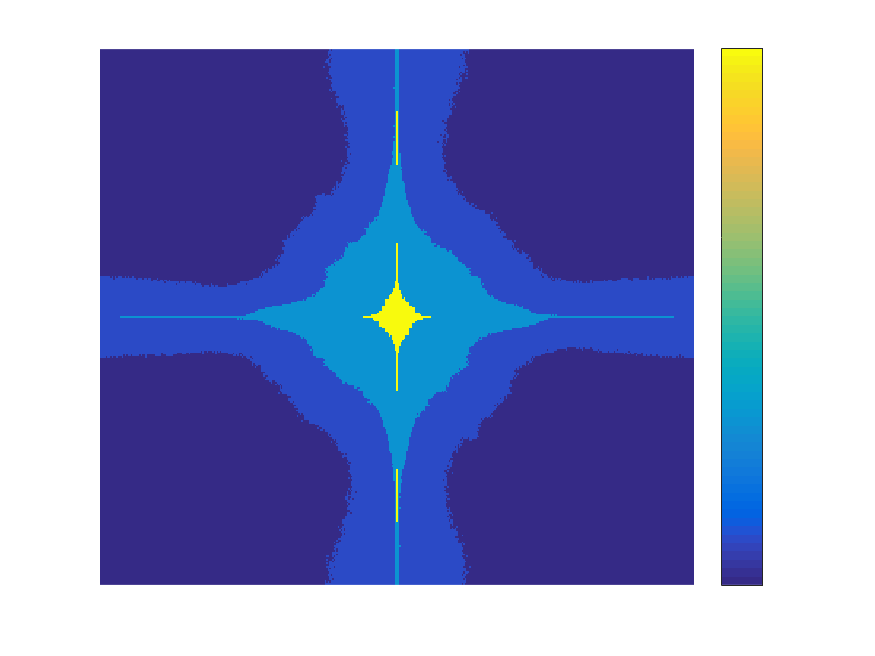}}\\
    \rowname{Codec er.}
        {\includegraphics[width=\tempwidth]{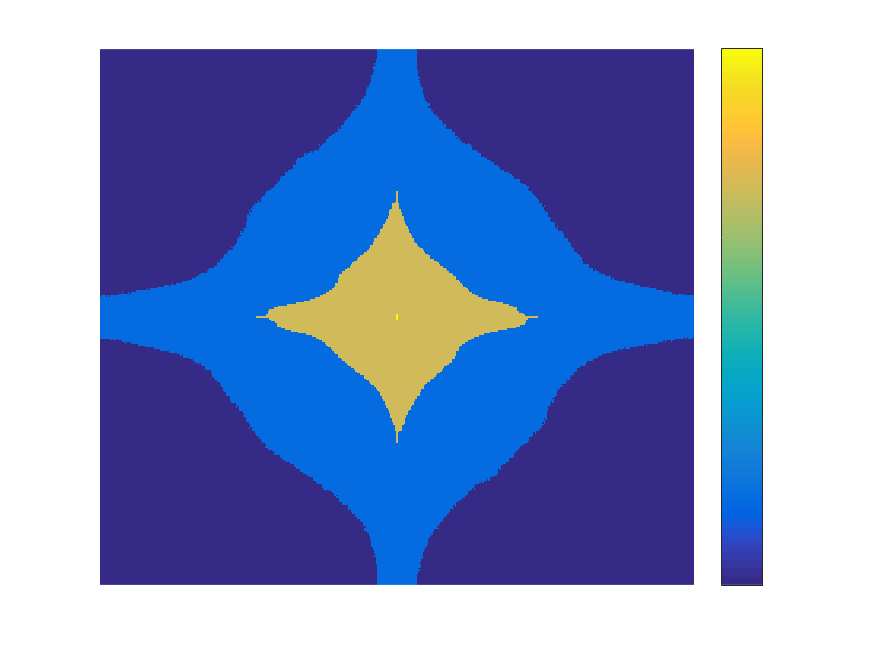}}\vspace{\vblank}\hfill
        {\includegraphics[width=\tempwidth]{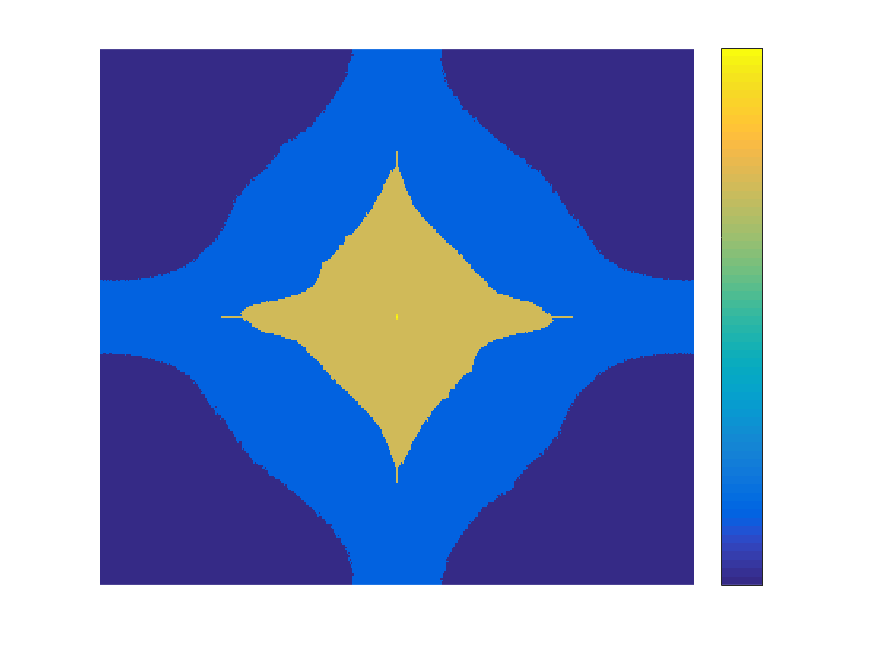}}\hfill
        {\includegraphics[width=\tempwidth]{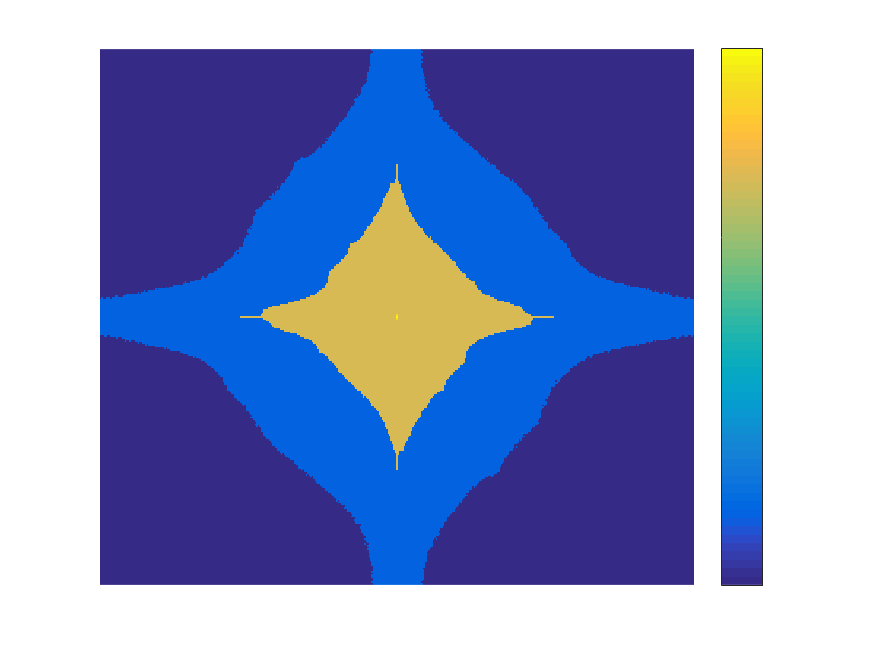}}\\
    \rowname{Darkening}
        {\includegraphics[width=\tempwidth]{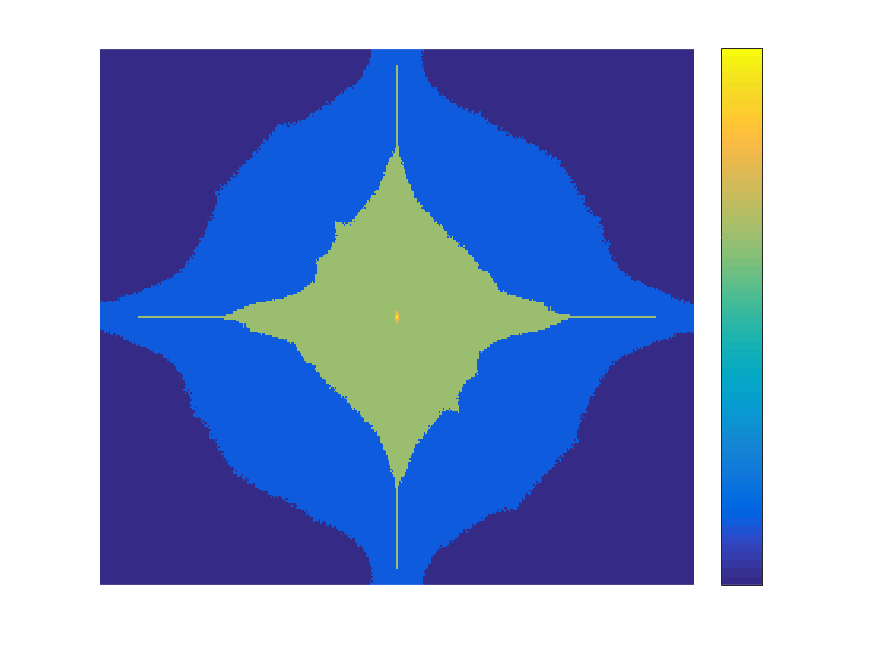}}\vspace{\vblank}\hfill
        {\includegraphics[width=\tempwidth]{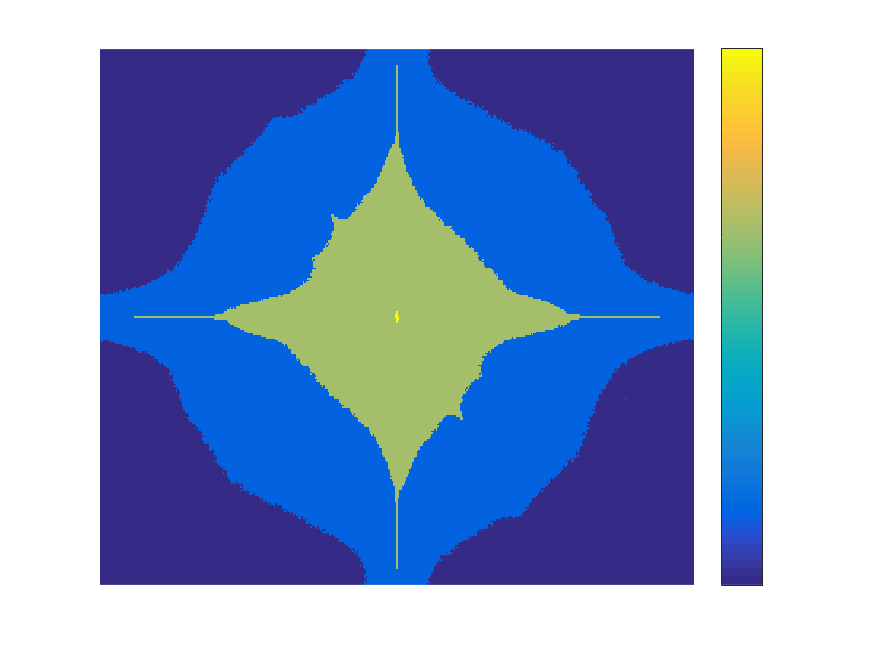}}\hfill
        {\includegraphics[width=\tempwidth]{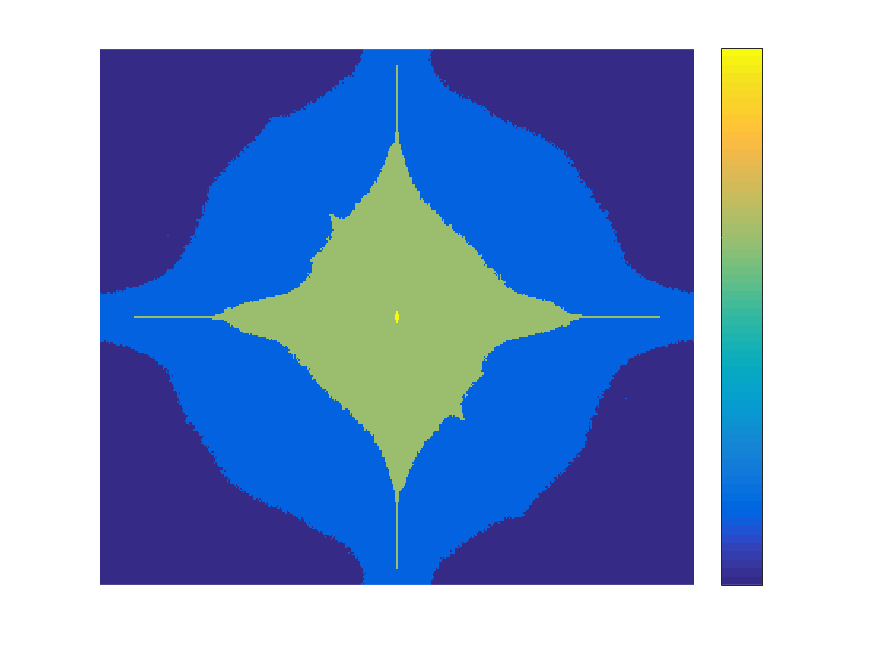}}\\
    \rowname{Dirty lens}
        {\includegraphics[width=\tempwidth]{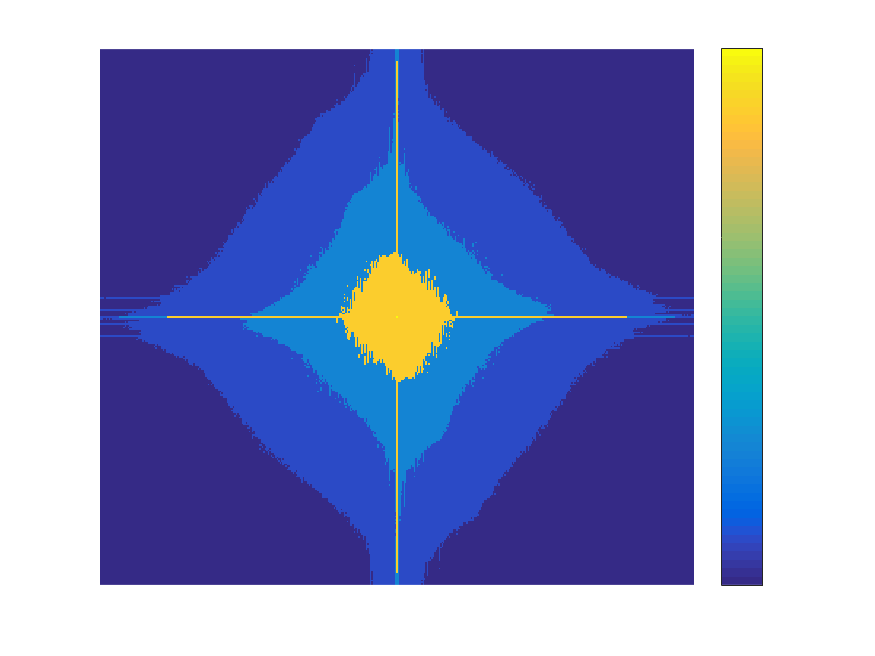}}\vspace{\vblank}\hfill
        {\includegraphics[width=\tempwidth]{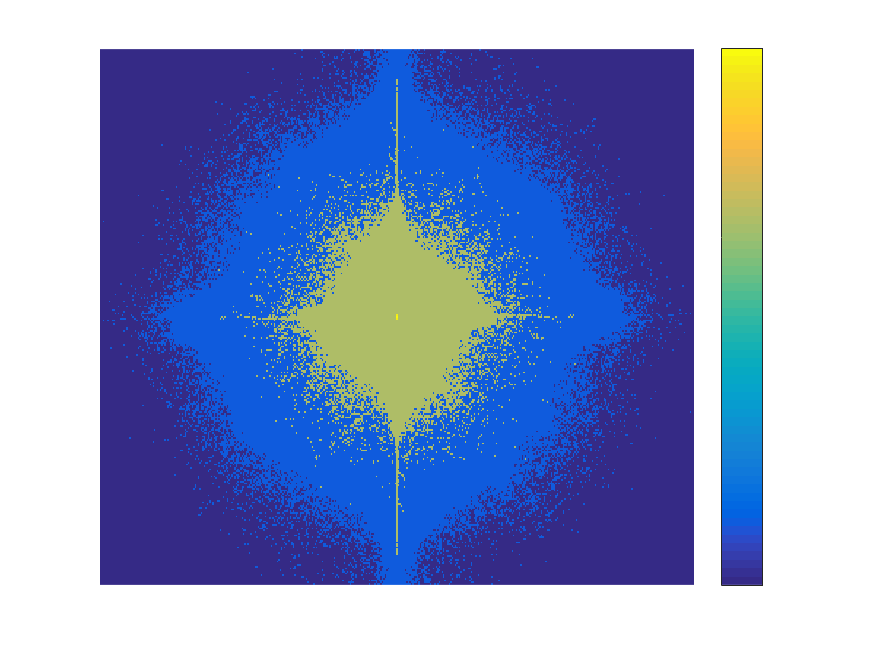}}\hfill
        {\includegraphics[width=\tempwidth]{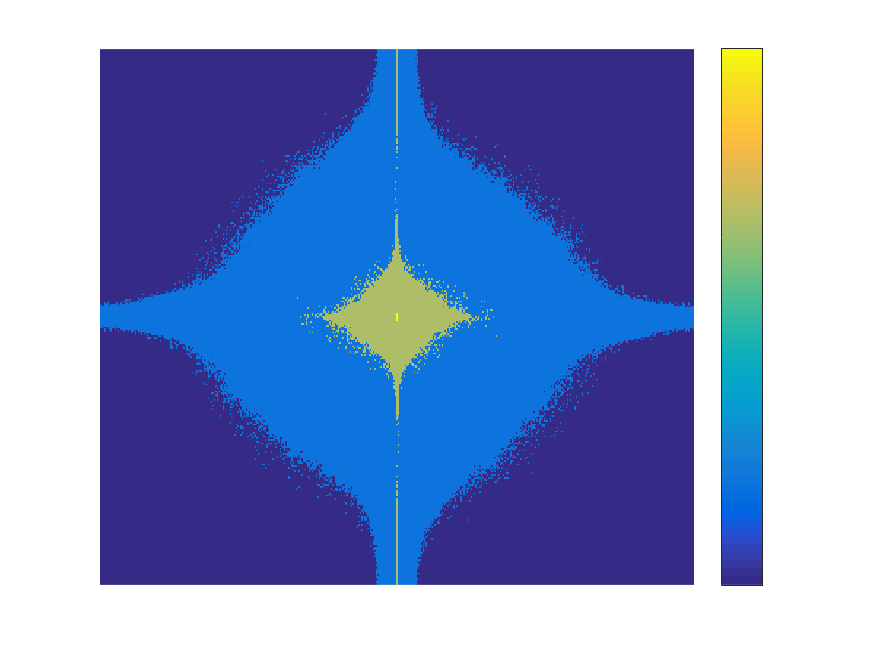}}\\
    \rowname{Exposure}
        {\includegraphics[width=\tempwidth]{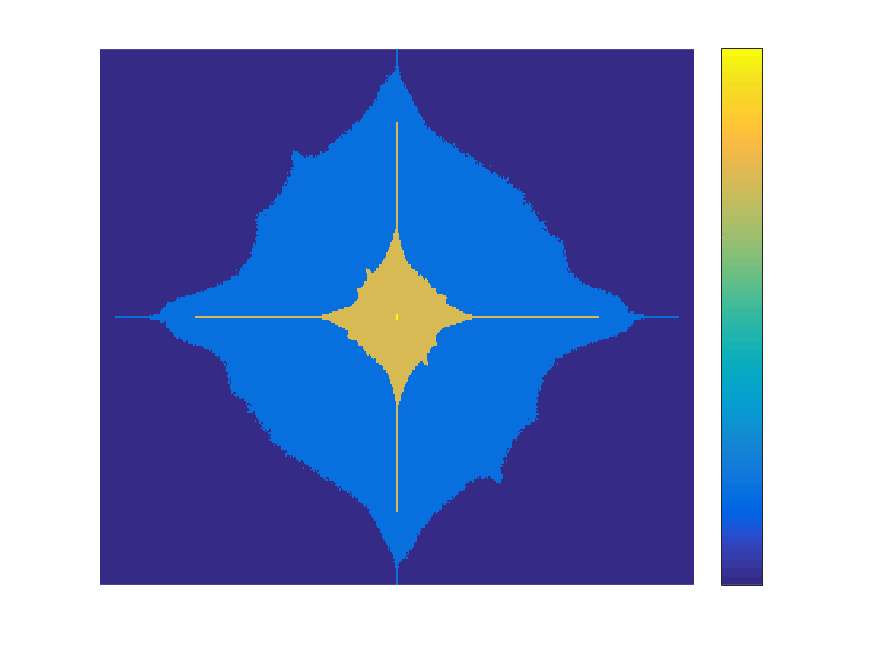}}\vspace{\vblank}\hfill
        {\includegraphics[width=\tempwidth]{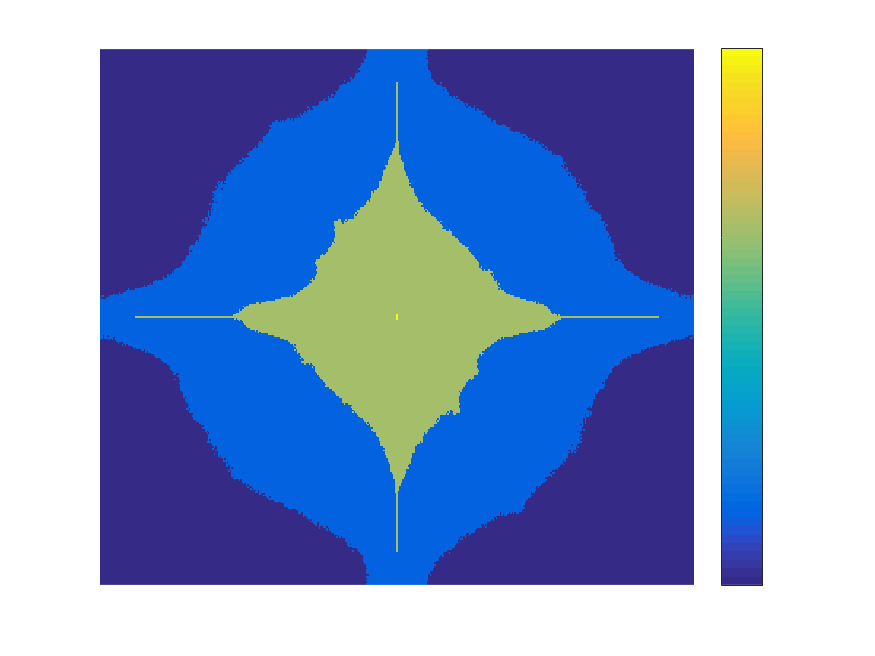}}\hfill
        {\includegraphics[width=\tempwidth]{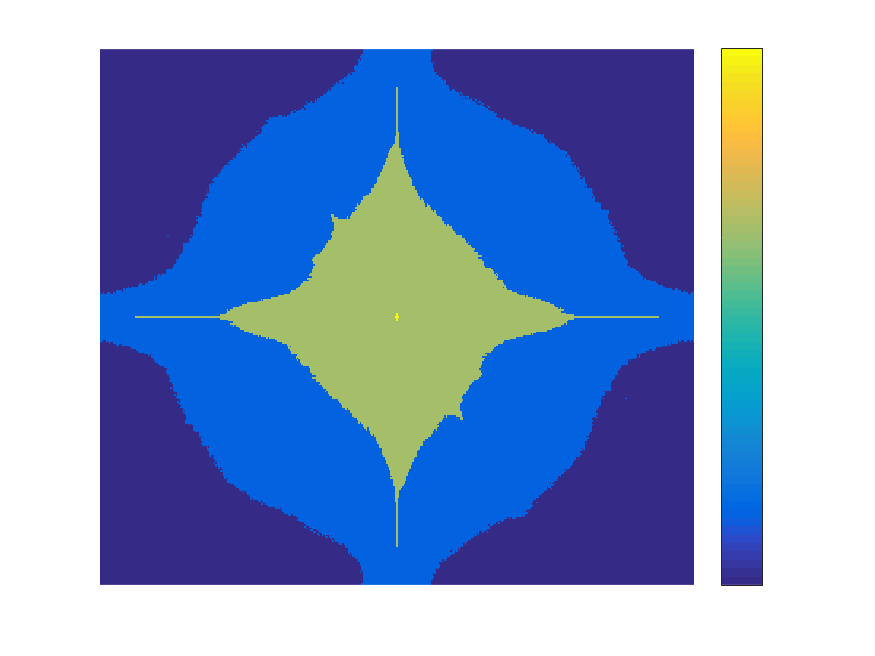}}\\
    \rowname{Gaus. blur}
        {\includegraphics[width=\tempwidth]{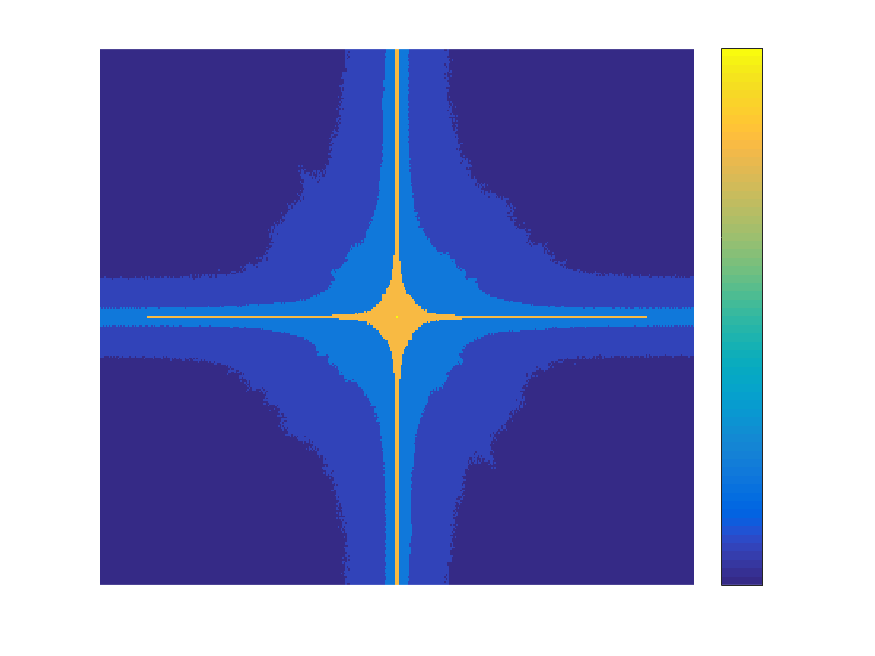}}\vspace{\vblank}\hfill
        {\includegraphics[width=\tempwidth]{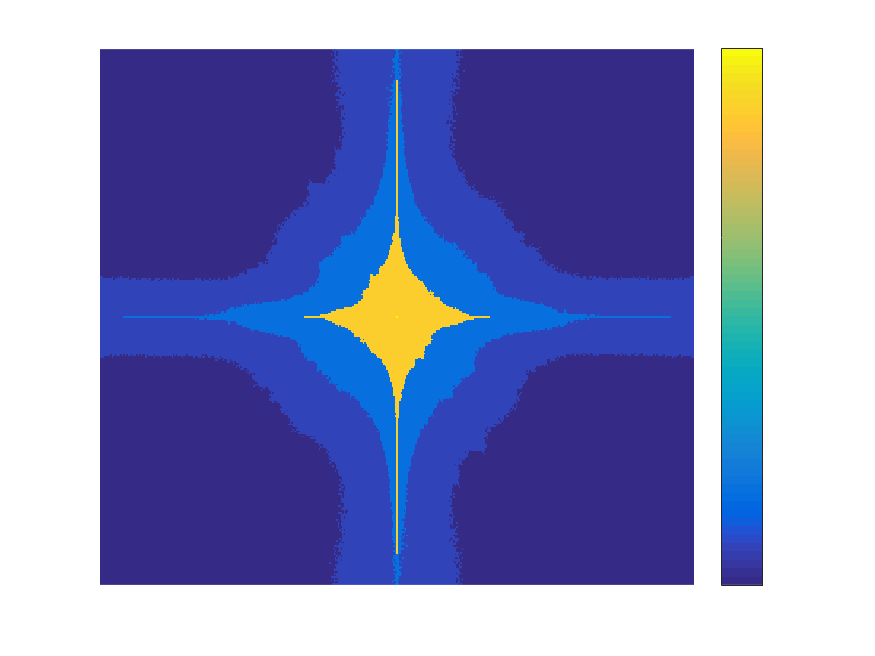}}\hfill
        {\includegraphics[width=\tempwidth]{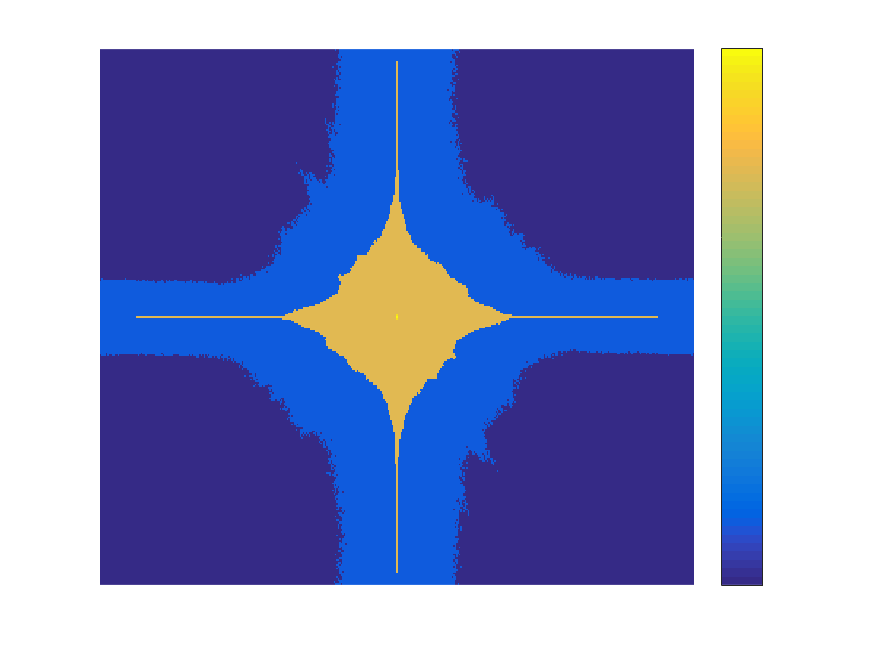}}\\
    \rowname{Noise}
        {\includegraphics[width=\tempwidth]{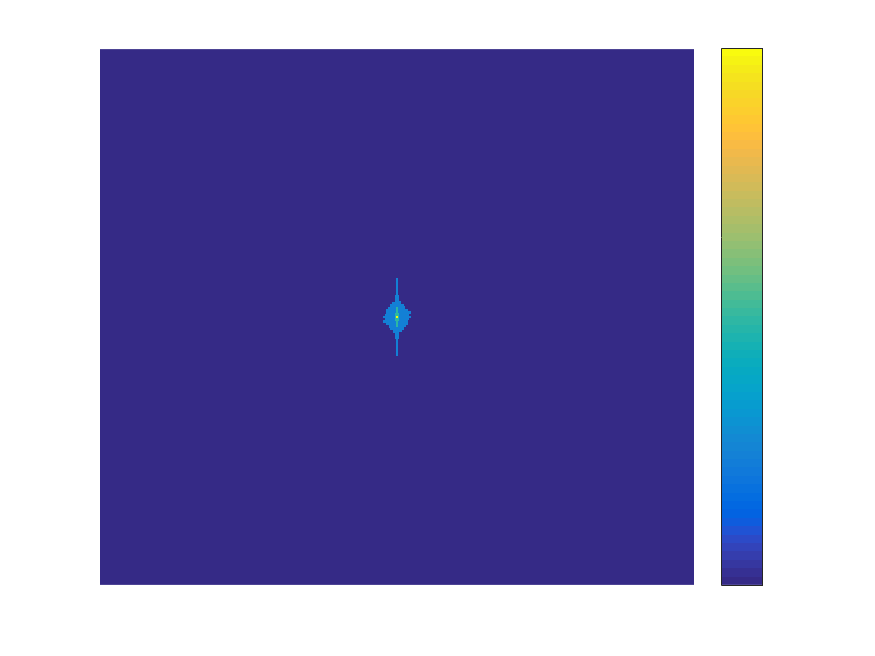}}\vspace{\vblank}\hfill
        {\includegraphics[width=\tempwidth]{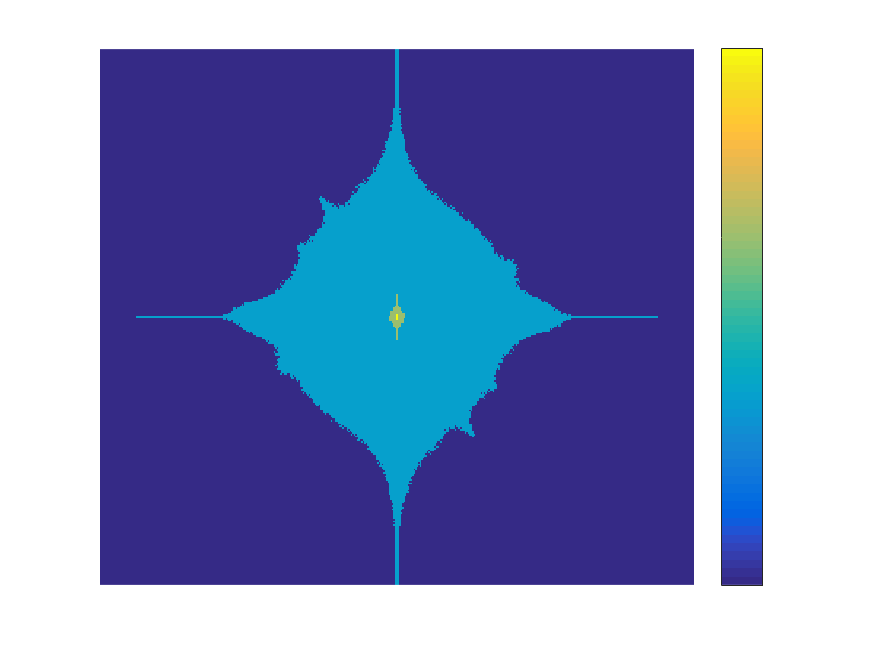}}\hfill
        {\includegraphics[width=\tempwidth]{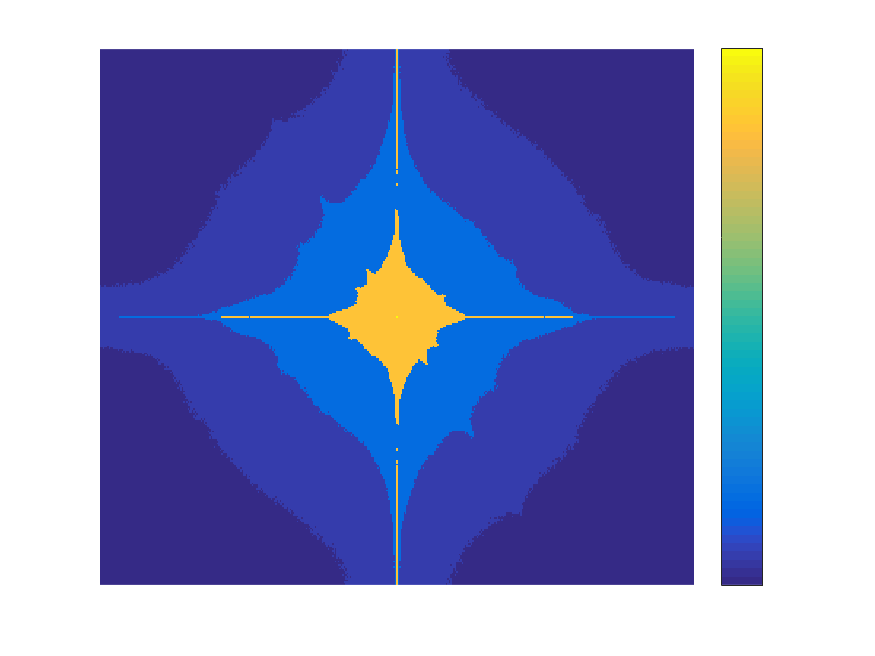}}\\
    \rowname{Rain}
        {\includegraphics[width=\tempwidth]{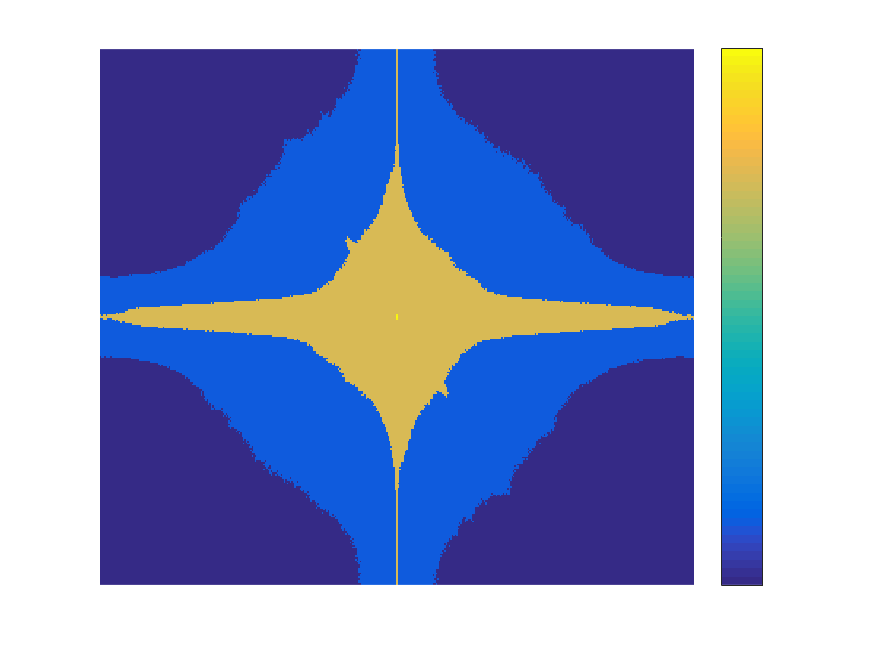}}\vspace{\vblank}\hfill
        {\includegraphics[width=\tempwidth]{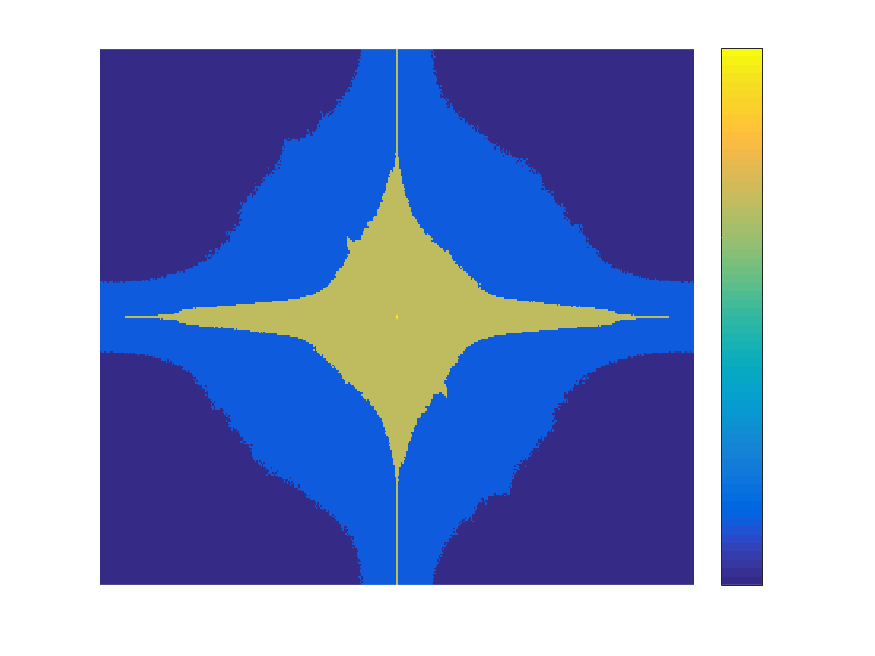}}\hfill
        {\includegraphics[width=\tempwidth]{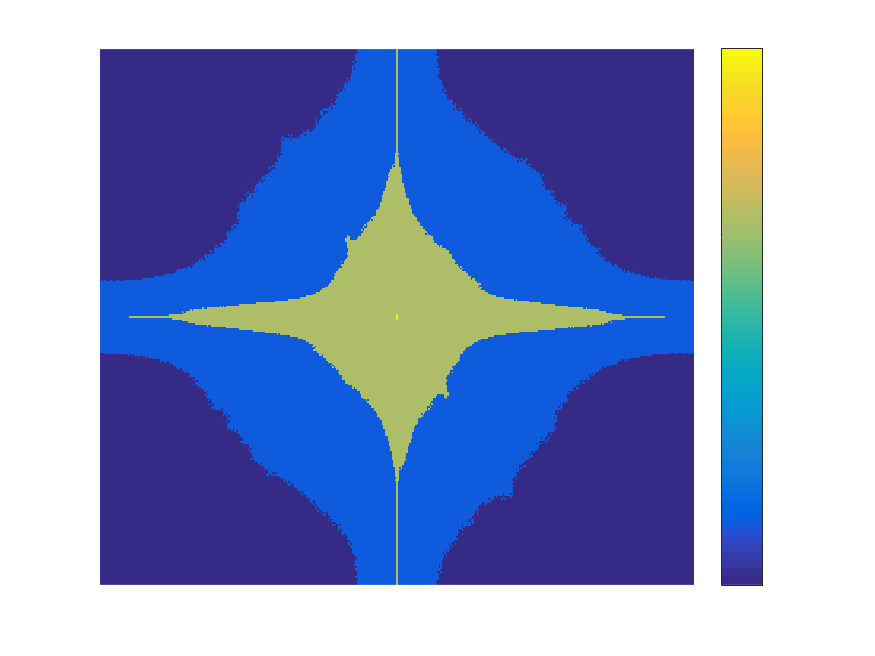}}\\
    \rowname{Shadow}
        {\includegraphics[width=\tempwidth]{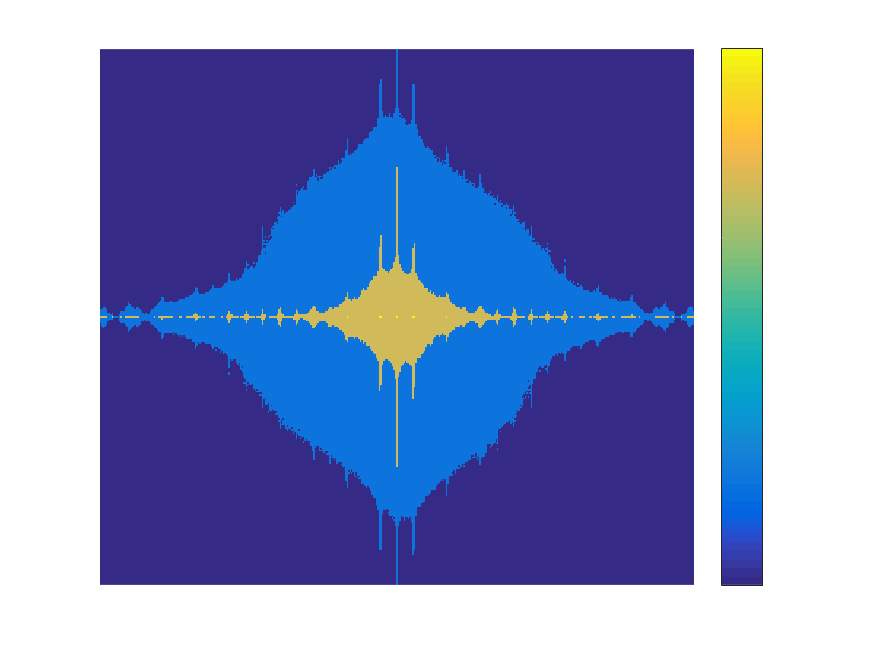}}\vspace{\vblank}\hfill
        {\includegraphics[width=\tempwidth]{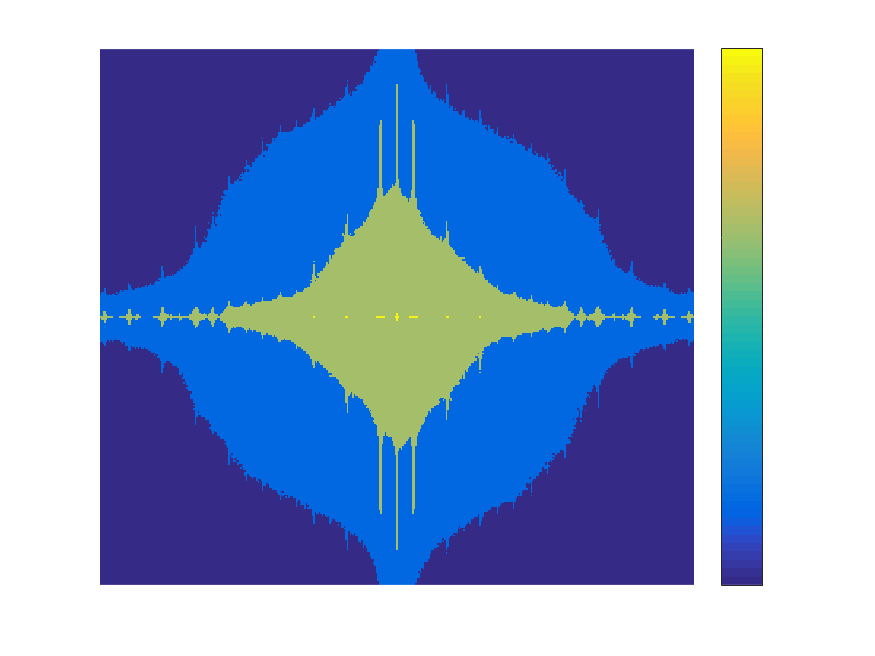}}\hfill
        {\includegraphics[width=\tempwidth]{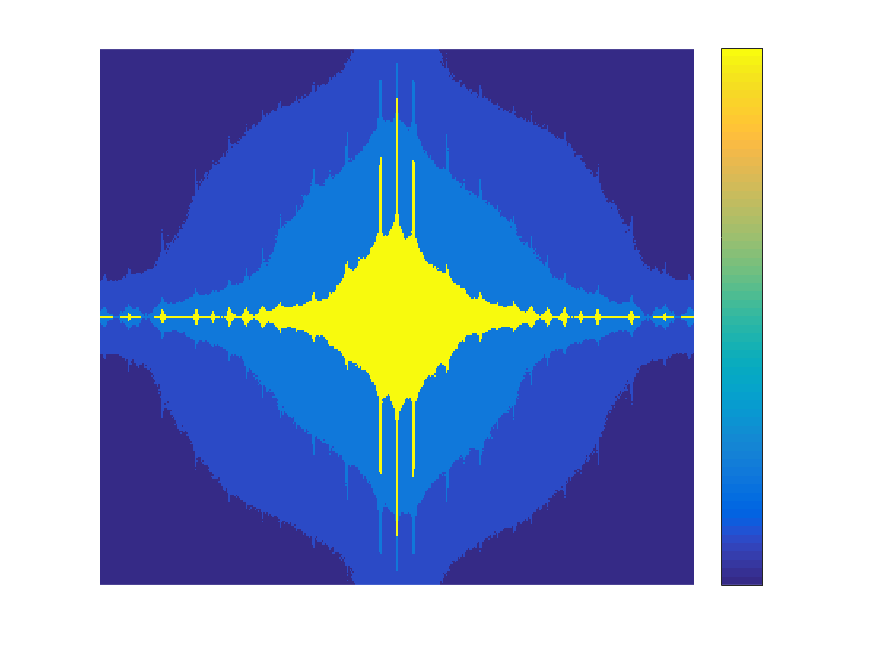}}\\
    \rowname{snow}
        {\includegraphics[width=\tempwidth]{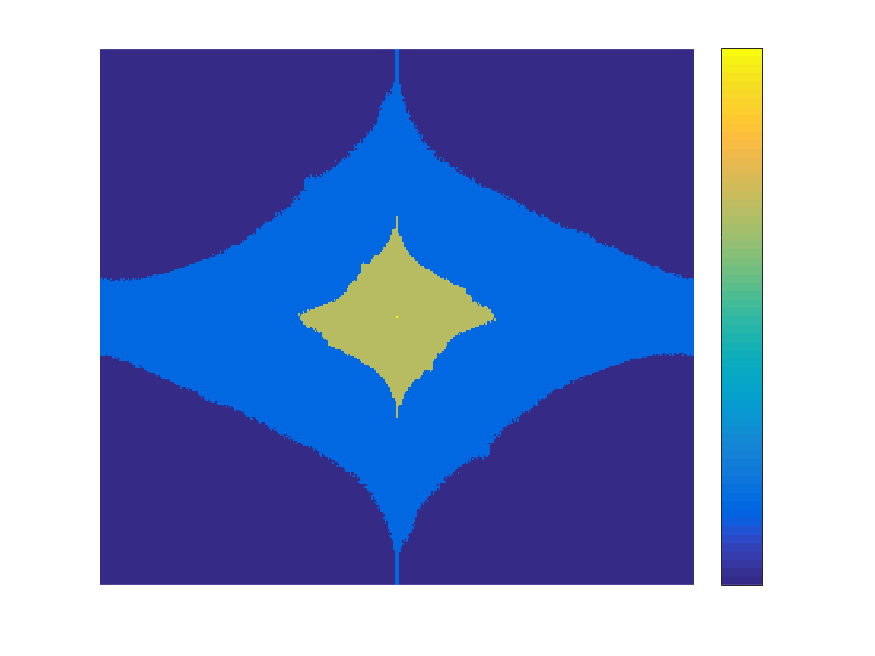}}\vspace{\vblank}\hfill
        {\includegraphics[width=\tempwidth]{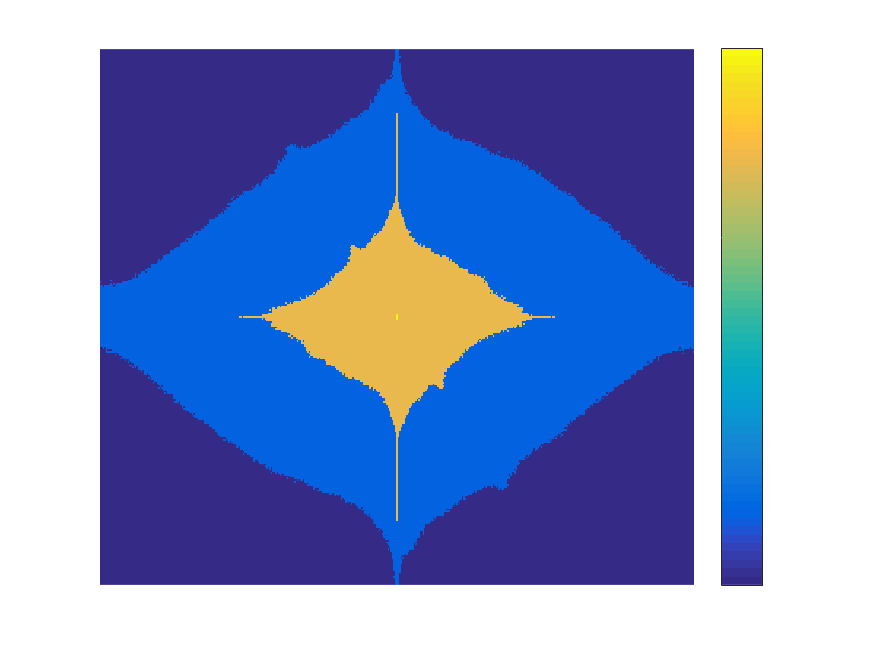}}\hfill
        {\includegraphics[width=\tempwidth]{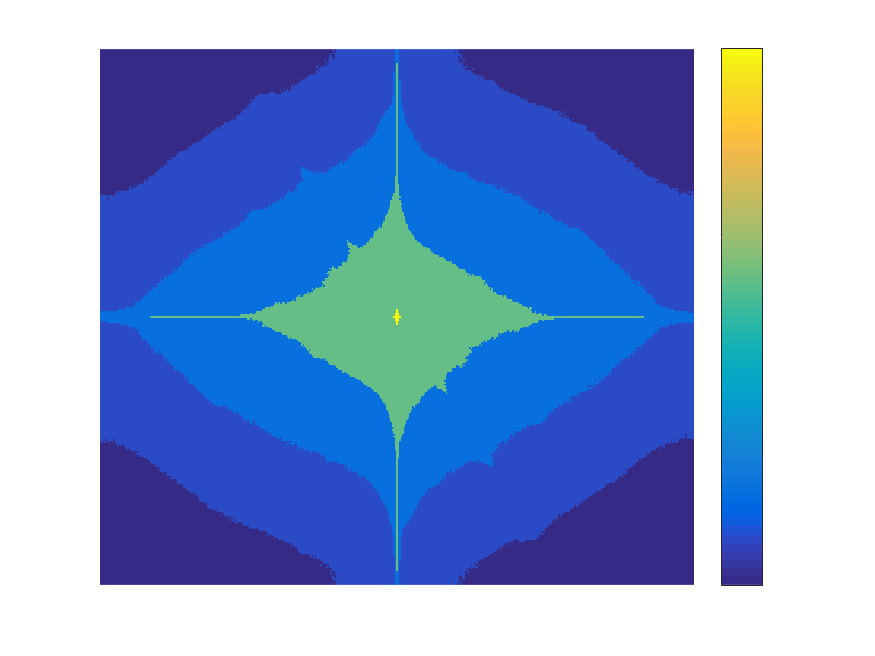}}\\
    \rowname{Haze}
        {\includegraphics[width=\tempwidth]{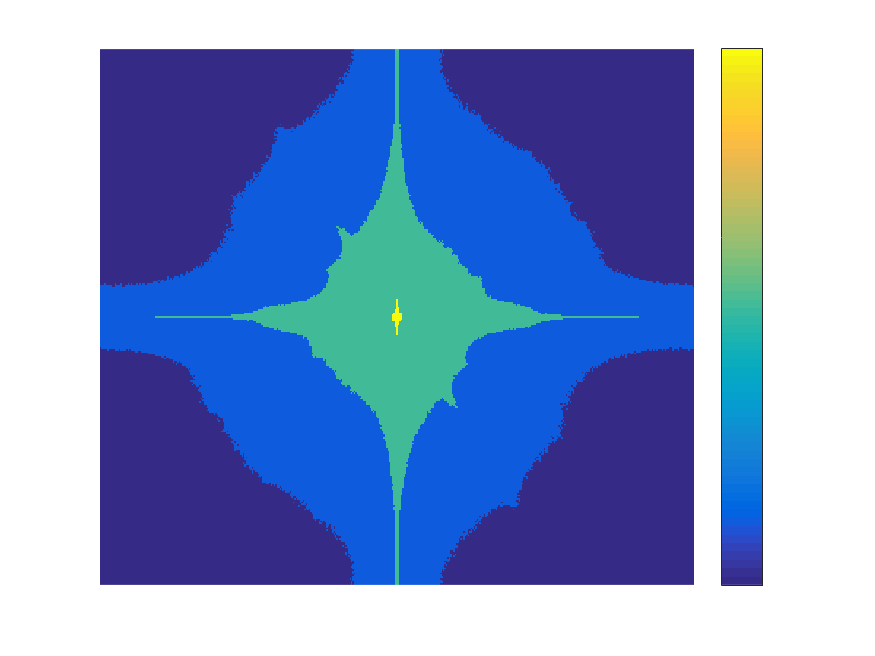}}\vspace{\vblank}\hfill
        {\includegraphics[width=\tempwidth]{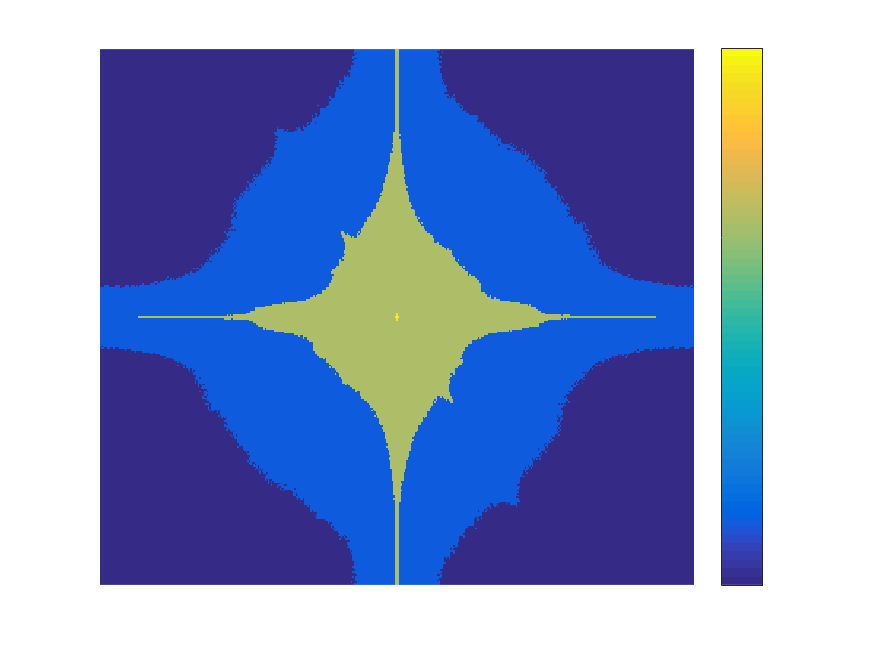}}\hfill
        {\includegraphics[width=\tempwidth]{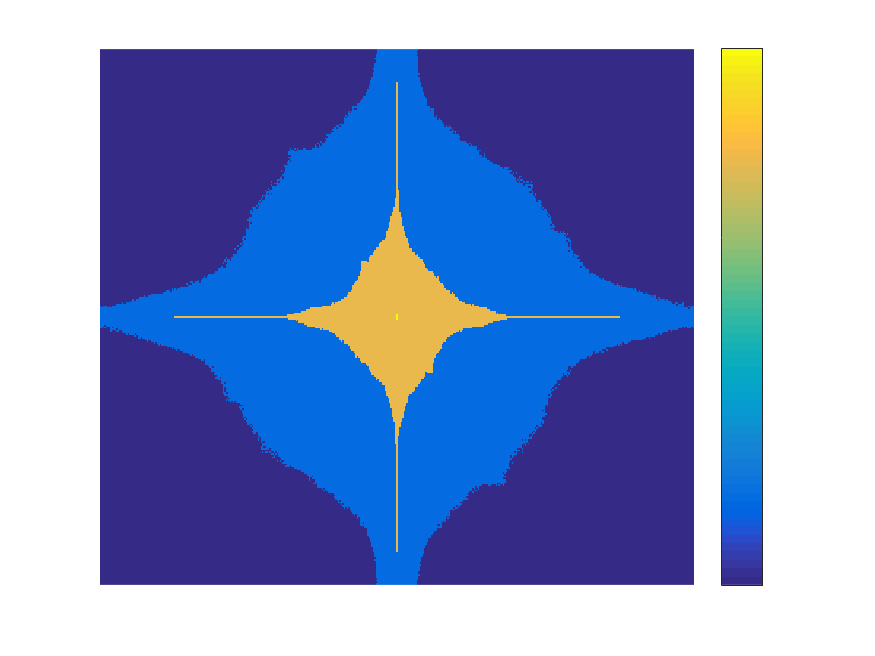}}\\
         \vspace{-5.0mm}
\caption{Average magnitude spectrum maps of video sequences for different challenge levels and types. }
\label{fig:spectrum_levels}

\end{figure}

Previously, we investigated the effect of challenge types in spectral representations of residual videos. To understand the effect of challenge levels in spectral representations, we also need to analyze the changes in spectrum with respect to the severity of challenging conditions. In Fig.~\ref{fig:spectrum_levels}, we show the average magnitude spectrum maps of video sequences for different challenge levels and types. Minor condition corresponds to level one, medium condition corresponds to level three, and major condition corresponds to level five challenges. Each spectrum corresponds to the average of $49$ video sequences with a specific challenge type and level. The most significant spectral change with respect to challenge levels occur in case of $decolorization$ and $noise$  whereas the least change occur in case of $codec~error$, $darkening$, and $rain$.

\begin{figure}[htbp!]
\begin{minipage}[b]{0.32\linewidth}
  \centering
\includegraphics[width=\linewidth, trim= 8mm 10mm 18mm 8mm]{Figs/mapsLevels/type6_level1.png}
  \vspace{0.03cm}
  \centerline{\scriptsize{(a)
      {\tabular[t]{@{}l@{}} Minor exposure  \\~ \endtabular}}}
\end{minipage}
 \vspace{0.05cm}
\begin{minipage}[b]{0.32\linewidth}
  \centering
\includegraphics[width=\linewidth, trim= 8mm 10mm 18mm 8mm]{Figs/mapsLevels/type6_level5.png}
  \vspace{0.03 cm}
  \centerline{\scriptsize{(b)
      {\tabular[t]{@{}l@{}} Major exposure  \\~ \endtabular}}}
\end{minipage}
 \vspace{0.05cm}
\begin{minipage}[b]{0.32\linewidth}
  \centering
\includegraphics[width=\linewidth, trim= 8mm 10mm 18mm 8mm]{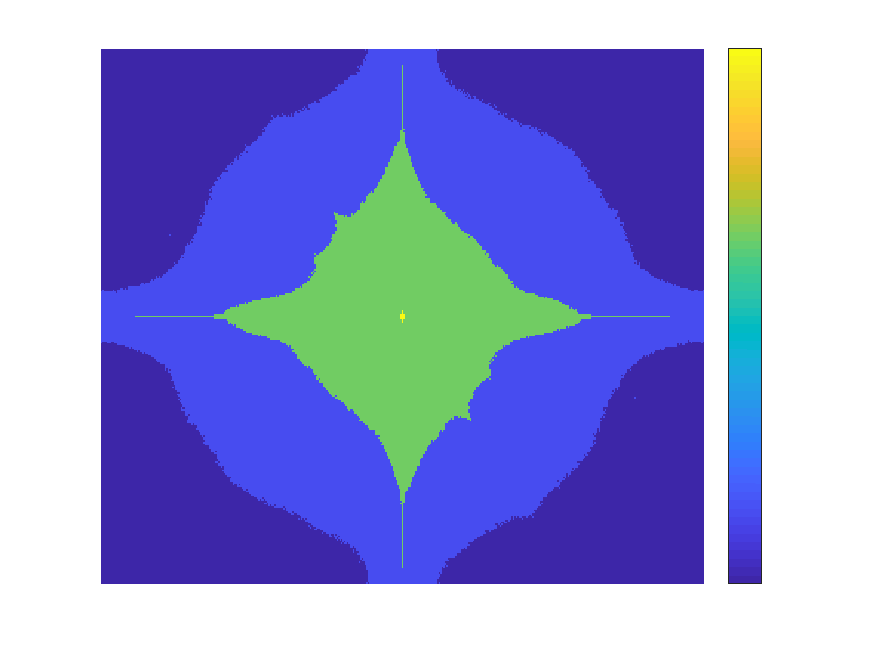}
  \vspace{0.03 cm}
  \centerline{\scriptsize{(c)
      {\tabular[t]{@{}l@{}} Major exposure \\and minor rain  \endtabular}}}
\end{minipage}
 \vspace{0.05cm}
\begin{minipage}[b]{0.32\linewidth}
  \centering
\includegraphics[width=\linewidth, trim= 8mm 10mm 18mm 4mm]{Figs/mapsLevels/type9_level1.png}
  \vspace{0.03cm}
  \centerline{\scriptsize{(d)
      {\tabular[t]{@{}l@{}} Minor rain  \\~ \endtabular}}}
\end{minipage}
 \vspace{0.05cm}
\begin{minipage}[b]{0.32\linewidth}
  \centering
\includegraphics[width=\linewidth, trim= 8mm 10mm 18mm 4mm]{Figs/mapsLevels/type9_level1.png}
  \vspace{0.03cm}
  \centerline{\scriptsize{(e)
      {\tabular[t]{@{}l@{}} Major rain  \\~ \endtabular}}}
\end{minipage}
 \vspace{0.05cm}
\begin{minipage}[b]{0.32\linewidth}
  \centering
\includegraphics[width=\linewidth, trim= 8mm 10mm 18mm 4mm]{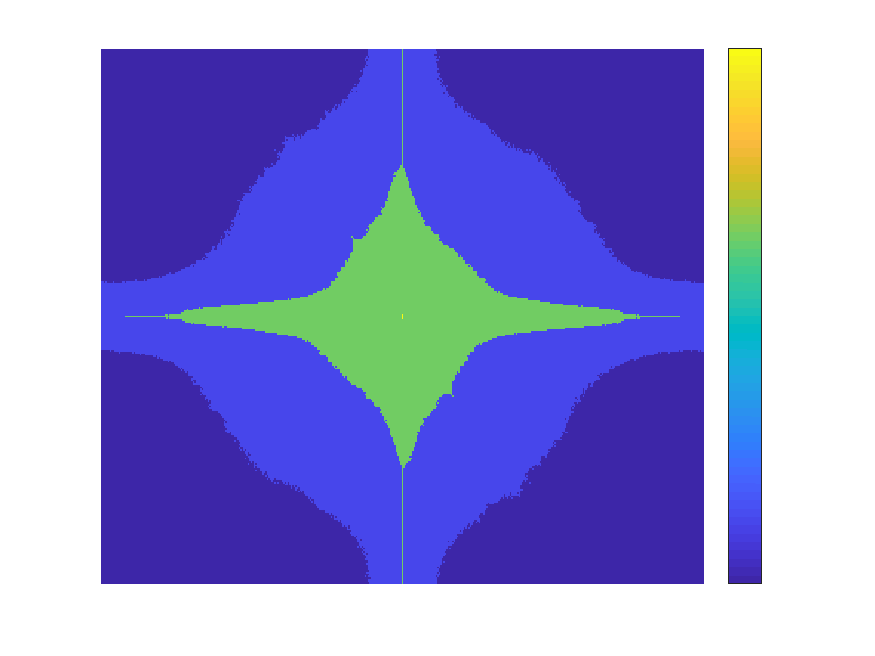}
  \vspace{0.03 cm}
  \centerline{\scriptsize{(f)
      {\tabular[t]{@{}l@{}} Major rain and  \\minor exposure \endtabular}}}
\end{minipage}
 \caption{Average magnitude spectrum maps of video sequences corresponding to rain, exposure, and a combination of rain and exposure at different challenge levels.}
\label{fig:spectrum_rain_expo}
\vspace{-4 mm}
\end{figure}

In the aforementioned experiments and analysis, we focused on the effect of individual challenging conditions and levels because video sequences in the \texttt{CURE-TSD-Real} dataset include one challenging condition at a time. Therefore, it is not possible to directly assess the effect of concurrent challenging conditions. In order to test the capability of spectral representations under concurrent challenging conditions with an example, we combined $rain$ and $exposure$ conditions and obtained their magnitude spectrums as shown in Fig.~\ref{fig:spectrum_rain_expo}. We obtained each magnitude spectrum map by averaging frame-level magnitude spectrums of 49 video sequences (14,700 frames). Spectral maps corresponding to concurrent $rain$ and $exposure$ conditions are shown in  Fig.~\ref{fig:spectrum_rain_expo}(c) and Fig.~\ref{fig:spectrum_rain_expo}(f). In addition to spectral maps of concurrent $rain$ and $exposure$ conditions, we included the spectral maps of isolated $rain$ and $exposure$ conditions in Fig.~\ref{fig:spectrum_rain_expo}(a), Fig.~\ref{fig:spectrum_rain_expo}(b), Fig.~\ref{fig:spectrum_rain_expo}(d), and Fig.~\ref{fig:spectrum_rain_expo}(e) to visually compare them next to each other.

In case of concurrent conditions, we can observe that major condition dominates the spectral representation. Spectral map of concurrent major $rain$ and minor $exposure$ (Fig.~\ref{fig:spectrum_rain_expo}(f)) is similar to the spectral maps of $rain$ (Fig.~\ref{fig:spectrum_rain_expo}(d-e)) in terms of asymmetry between horizontal and vertical components.  Moreover, spectral map of concurrent major $exposure$ and minor $rain$ (Fig.~\ref{fig:spectrum_rain_expo}(c)) is similar to the spectral map of major exposure (Fig.~\ref{fig:spectrum_rain_expo}(b)). Thus, we can mention that dominant conditions mostly determine the shape of the spectral maps. However, we can still observe differences in the spectral maps when we compare concurrent major and minor condition with solely major condition. For example, spectral map of major $rain$ (Fig.~\ref{fig:spectrum_rain_expo}(e)) and spectral map of concurrent major $rain$ and minor $exposure$ (Fig.~\ref{fig:spectrum_rain_expo}(f)) are still separable from each other in terms of shape and color. Meanwhile, spectral map of major $exposure$ and minor $rain$ (Fig.~\ref{fig:spectrum_rain_expo}(c)) and spectral map of major $exposure$ (Fig.~\ref{fig:spectrum_rain_expo}(b)) are separable from each other in terms of color, which reflects differences in terms of spectral magnitude. Based on this example, we can express that spectral maps can reflect the impact of two concurrent conditions, but identification of the concurrent conditions may not be as straightforward as the identification of individual conditions.

\subsection{Detection Performance versus Spectral Characteristics}
\label{subsec:perf_estimation}
\begin{figure}[htbp!]
    \centering
    \includegraphics[width=0.8\linewidth, trim= 40mm 85mm 40mm 88mm
]{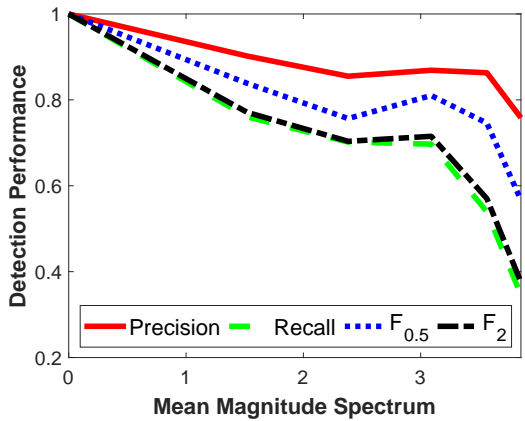}
    \caption{Detection performance versus mean magnitude spectrum of residual video sequences.}
    \label{fig:perf_est}
    \vspace{-4 mm}
\end{figure}

Even though challenge levels affect the spectral representations, high level spectral shapes remain similar in majority of the challenging conditions. The intensity of the magnitude spectrums can be used to quantify the changes in spectral representations, which can be an indicator of the detection performance degradations. In Fig.~\ref{fig:perf_est}, we show the relationship between detection performance and mean magnitude spectrum. Specifically, we computed the detection performance under varying challenge levels and calculated the mean magnitude spectrum corresponding to the varying challenge levels. We can observe that an increase in mean magnitude spectrum generally corresponds to a decrease in detection performance. To measure the correlation between traffic sign detection performance and mean magnitude spectrum, we calculated the Spearman rank order correlation coefficient, which is reported for each detection performance metric in Table \ref{tab:perf_est}. Specifically, we measured the correlation between mean magnitude spectrum and detection performance for each challenge category and obtained the average of these correlation coefficients. Based on the experiments, correlation between detection performance and mean magnitude spectrum is $0.643$ for precision, $0.848$ for recall, $0.657$ for $F_{0.5}$ score, and $0.810$ for $F_{2}$ score.

Spectral representations can be used to analyze the changes in images and videos and these changes can be quantified by measuring the changes in spectral representations. A direct mean pooling operation is a straightforward approach to quantify spectrums of residual sequences. However, detection algorithms do not necessarily react identically to changes at different frequencies. Therefore, instead of a direct mean pooling operation, a weighted pooling can be performed by considering the relative importance of frequency bands for traffic sign detection. For example, in JPEG compression \cite{Wallace1992}, the objective is to compress the image as much as possible without visual artifacts. To achieve this objective, quantization tables were designed based on psychovisual experiments to compress signal components according to the their perceptual significance. Similarly, a significance map can be designed for traffic sign detection application to quantify the changes in spectral components according to their algorithmic significance. Spectral analysis approach investigated in this study requires a reference video. Therefore, to estimate the traffic sign detection performance, we need to obtain the images of the same scene at different conditions. Such a system is feasible for a fixed camera setup in which we can capture the same region at different times. To deploy such systems to mobile platforms, we need to focus on no-reference spectral representations in which there is no need for a reference video.

\begin{table}[htbp!]
\small
\centering
\caption{Detection performance degradation estimation with mean magnitude spectrum under challenging conditions.}
\label{tab:perf_est}
\begin{tabular}{ccccc}
\hline
\textbf{Estimated Metric}                              & \textbf{Precision}      & \textbf{Recall}          & $\mathbf{F_{0.5}}$      & $\mathbf{F_{2}}$                                    \\ \hline
\bf \begin{tabular}[p]{@{}c@{}} Estimation Performance\\ (Spearman Correlation)  \end{tabular} & 0.643 & 0.848 & 0.657 & 0.810 
\\ \hline \hline
\vspace{-4 mm}
\end{tabular}
\end{table}

\section{Conclusion}
\label{sec:conc}
We analyzed the average performance of benchmark algorithms in the \texttt{CURE-TSD-Real} dataset and showed that detection performance can significantly degrade under challenging conditions. $Codec~error$ and $exposure$ resulted in the most significant performance degradation with more than $80\%$ whereas $shadow$ resulted in the least degradation with around $16\%$. Challenging weather conditions $snow$, $haze$, and $rain$ resulted in at least $48\%$ performance degradation.  Around $63\%$ performance degradation in $decolorization$ highlighted the importance of color information for certain algorithms in sign detection. Detection performance degradation based on $darkening$, $noise$ and $blur$ is in between $30\%$ and $48\%$ whereas $dirty~lens$ exceeds $50\%$. Our frequency domain analysis showed that simulated challenging conditions can correspond to distinct spectral patterns and magnitude of these spectral patterns can be used to estimate the detection performance under challenging conditions. Degradation estimation perfomance based on spectral representations was in between $0.64$ and $0.84$ in terms of Spearmen correlation. As future work, adaptive pooling and no-reference spectral analysis are promising research directions that can be further investigated to estimate detection performance of algorithms by solely considering the environmental conditions.

\ifCLASSOPTIONcaptionsoff
  \newpage
\fi


\vspace{-12.0mm}

\begin{IEEEbiography}
    [{\includegraphics[width=1.0in]{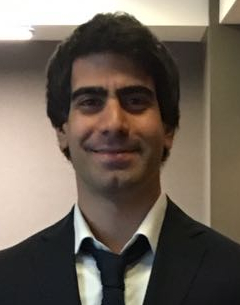}}]{Dogancan Temel} received his
M.S. and Ph.D. degrees in electrical and computer engineering (ECE) from the Georgia
Institute of Technology (Georgia Tech) in 2013 and 2016, respectively, where he is currently a postdoctoral fellow. During his PhD, he received Texas Instruments Leadership Universities Fellowship for four consecutive years. He was the recipient of the Best Doctoral Dissertation Award from the Sigma Xi honor society, the Graduate Research Assistant Excellence Award from the School of ECE, and the Outstanding Research Award from the Center for Signal and Information Processing at Georgia Tech. His research interests are focused on human and machine vision.

\end{IEEEbiography}

\vspace{-18.0mm}

\begin{IEEEbiography}
    [{\includegraphics[width=1in]{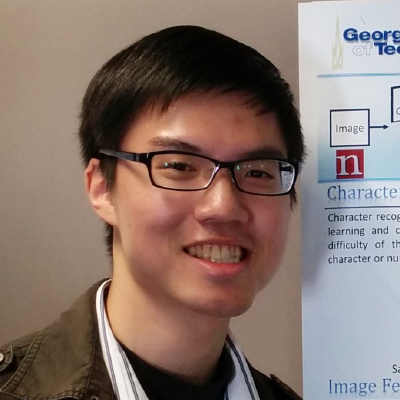}}]{Min-Hung Chen}  received an M.S. degree in 2012 from National Taiwan University. He worked as a research assistant in Academia Sinica, Taiwan in 2013. He is currently working towards PhD in the school of Electrical and Computer Engineering in Georgia Institute of Technology, Atlanta. During his PhD, he received Ministry of Education Technologies Incubation Scholarship from Taiwan for three years. His research interests include computer vision, deep learning, image and video processing. Recently, Min-Hung Chen is investigating the temporal dynamics through deep learning techniques.
\end{IEEEbiography}

\vspace{-12.0mm}

\begin{IEEEbiography}
    [{\includegraphics[width=1in]{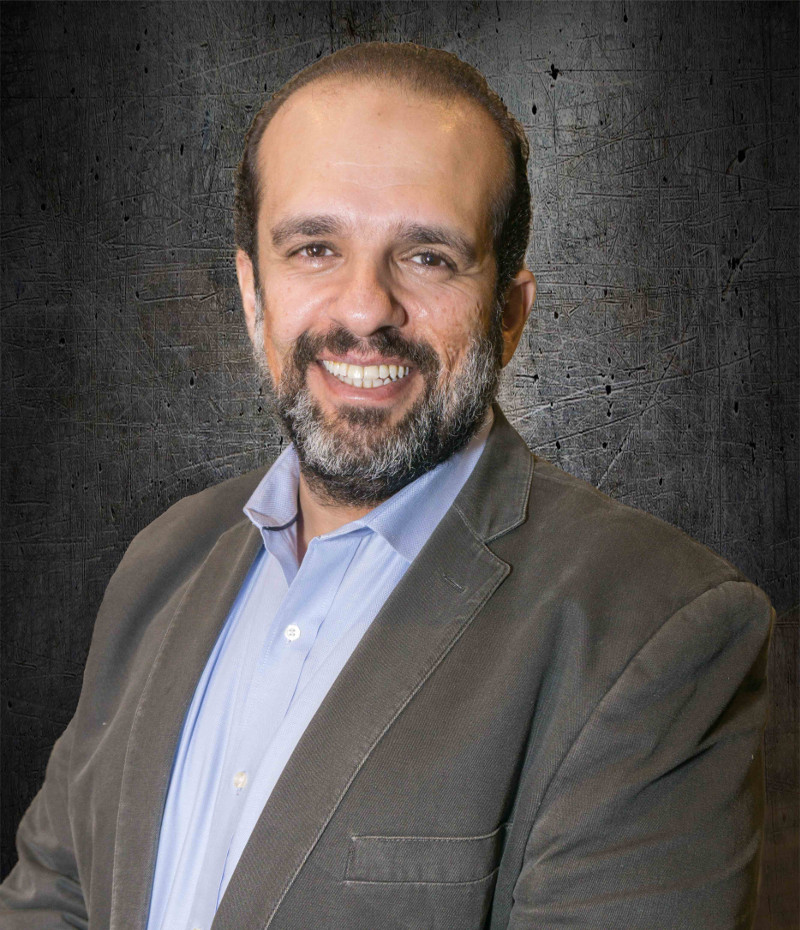}}]{Ghassan AlRegib} is currently a Professor with the School of Electrical and Computer Engineering, Georgia Institute of Technology.  He was a recipient of the ECE Outstanding Graduate Teaching Award in 2001 and both the CSIP Research and the CSIP Service Awards in 2003, the ECE Outstanding Junior Faculty Member Award, in 2008, and the 2017 Denning Faculty Award for Global Engagement. His research group, the Omni Lab for Intelligent Visual Engineering  and Science (OLIVES) works on research projects related to machine learning, image and  video processing, image and video understanding, seismic interpretation, healthcare intelligence, machine learning for ophthalmology, and video analytics. He participated in several service activities within the IEEE including the organization of the First IEEE VIP Cup (2017), Area Editor for the IEEE Signal Processing Magazine, and the Technical Program Chair of GlobalSIP’14 and ICIP’20. He has provided services and consultation to several firms, companies, and international educational and Research and Development organizations. He has been a witness expert in a number of patents infringement cases. 
\end{IEEEbiography}

\end{document}